\documentclass{SCIS2023}

\newcommand{\RNum}[1]{\uppercase\expandafter{\romannumeral #1\relax}}
\usepackage{multirow}
\usepackage{colortbl}
\usepackage{soul}
\usepackage{color}
\definecolor{hl}{rgb}{0.75,0.75,0.75}
\sethlcolor{hl}
\usepackage{graphicx}

\begin{document}
\ArticleType{RESEARCH PAPER}
\Year{2022}
\Month{}
\Vol{}
\No{}
\DOI{}
\ArtNo{}
\ReceiveDate{}
\ReviseDate{}
\AcceptDate{}
\OnlineDate{}

\title{Enhancing SAEAs with Unevaluated Solutions: \\A Case Study of Relation Model for Expensive Optimization}{}

\author[1]{Hao Hao}{}
\author[1]{Xiaoqun Zhang}{}
\author[2,3]{Aimin Zhou}{amzhou@cs.ecnu.edu.cn}

\AuthorMark{Hao Hao}

\AuthorCitation{Hao Hao, Xiaoqun Zhang, Aimin Zhou}


\address[1]{Institute of Natural Sciences, Shanghai Jiao Tong University, Shanghai {\rm 200240}, China}
\address[2]{Shanghai Frontiers Science Center of Molecule Intelligent Syntheses, Shanghai {\rm 200062}, China}
\address[3]{School of Computer Science and Technology, East China Normal University, Shanghai {\rm 200062}, China}

\abstract{
Surrogate-assisted evolutionary algorithms~(SAEAs) hold significant importance in resolving expensive optimization problems~(EOPs). Extensive efforts have been devoted to improving the efficacy of SAEAs through the development of proficient model-assisted selection methods. However, generating high-quality solutions is a prerequisite for selection. The fundamental paradigm of evaluating a limited number of solutions in each generation within SAEAs reduces the variance of adjacent populations, thus impacting the quality of offspring solutions. This is a frequently encountered issue, yet it has not gained widespread attention. This paper presents a framework using unevaluated solutions to enhance the efficiency of SAEAs. The surrogate model is employed to identify high-quality solutions for direct generation of new solutions without evaluation. To ensure dependable selection, we have introduced two tailored relation models for the selection of the optimal solution and the unevaluated population. A comprehensive experimental analysis is performed on two test suites, which showcases the superiority of the relation model over regression and classification models in the selection phase. Furthermore, the surrogate-selected unevaluated solutions with high potential have been shown to significantly enhance the efficiency of the algorithm.
}


\maketitle

\section{Introduction}

\begin{figure}[htbp]
  \centering
  \subfloat{\includegraphics[width=0.8\textwidth,trim=0 0 0 0,clip]{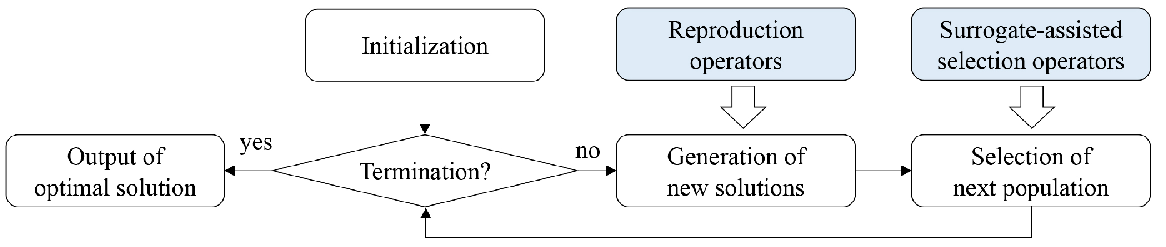}} \\
\caption{Flowchart of surrogate-assisted evolutionary algorithms.}
\label{fig:SAEA_framework}
\end{figure}

Valued for their global search capability and adaptability, evolutionary algorithms (EAs) are extensively utilized in various fields~\cite{back1997handbook}. Despite the common presumption of a method to assess each solution's fitness, expensive optimization problems often present significant challenges due to the extensive computational resources or costly experiments they require~\cite{jinSurrogateassistedEvolutionaryComputation2011a,DBLP:journals/chinaf/LiuHQQY22}. Addressing such practical limitations, surrogate-assisted evolutionary algorithms (SAEAs) have gained prominence. By integrating the robust global search of EAs with cost-effective surrogate model estimation, SAEAs have emerged as a mainstream method for solving these resource-intensive problems~\cite{LiEvolutionaryComputationExpensive2022}. The SAEA framework, as depicted in Figure~\ref{fig:SAEA_framework}, with `reproduction operators' and `surrogate-assisted selection operators' being the pivot around which the framework revolves. It is the duty of the reproduction operators to generate innovative trial solutions, thus expanding the exploration of the search space. Concurrently, the surrogate-assisted selection operators strategically select prospective high-quality solutions for real fitness evaluation. These two operators alternate execution to drive the population toward the optimal solution region.

An important question is how the aforementioned two modules, `reproduction operators' and `surrogate-assisted selection operators', cooperate in SAEAs. Given the limited evaluation budget, the choice of which solutions and how many to real evaluate will impact SAEAs' preferences in terms of exploration and exploitation. Simultaneously, the decision of which solutions to use as parent solutions for generating new ones also dictates the SAEAs' search direction. The answer to this issue depends on the algorithm's selection strategy, which can also be referred to as the model management strategy~\cite{zhan2020expected,jinSurrogateassistedEvolutionaryComputation2011a}. There are two typical methods, `select $N$' and `select $1$'. In `select $N$'~\cite{zhang2018preselection,haoBinaryRelationLearning2020}, a large number of trial solutions are generated using the reproduction operator, which exceeds the population size ($N$). Then, a surrogate model is utilized to select $N$ solutions as offspring solutions for real evaluation. This methodology presents the benefit of augmenting the quantity of superior solutions in each generation and allowing entry into the following generation to promote population mobility. However, it aggravates the depletion of the real function evaluation budget. On the contrary, in the `select $1$' method~\cite{journals/tec/LiuZG14,chen2020surrogate}, the benefit is a noteworthy reduction in the real evaluation budget. However, ideally, only one solution in the subsequent generation populace will be altered. As a result, there is a possibility that the distribution of the population may not be substantially enriched, and newly generated solutions may remain confined to the current search area~(In Section~\ref{subsec:variance}, a simple visualization experiment will confirm this)~. The acquisition function is used to balance the search preferences and improve the diversity of the population to a certain extent. However, its ability to improve the current population distribution is limited. Therefore, when only one solution is sampled, it is difficult for the new solution in the next generation to escape the current local optimum.

The aforementioned `select $N$' and `select $1$' strategies both present unique advantages and challenges. This prompts the question: Can we devise a simple method that amalgamates the strengths of both `select $N$' and `select $1$' strategies? A method where, in the current population, the solution deemed best by the surrogate model is real evaluated, and the external archive and surrogate model are updated accordingly. At the same time, certain high-quality solutions identified by the model, without real evaluations, are chosen to directly contribute to the generation of solutions for the following iteration. Even though these solutions may not necessarily be optimal, their potential to surpass some of the parent solutions in quality is plausible. Implementing such a method would not escalate the algorithm's evaluation cost, but could augment the population's diversity and accelerate the algorithm's progression towards the optimal region.

The successful implementation of the aforementioned proposal is contingent upon a pivotal prerequisite of dependable prediction results from surrogate models. A variety of regression~\cite{journals/tec/SunJCDZ17,willmes2003comparing} and classification~\cite{li2019ibea,journals/tec/PanHTWZJ19} models can be employed to ascertain solution quality~\cite{haoBinaryRelationLearning2020}. Despite the significant contributions of existing models, our goal in this paper is to develop surrogate models that are better aligned with the specific needs of the problem at hand. Considering the accomplishments of widely-used regression and classification models, we believe there's still room to create even more reliable surrogate models. To that end, we introduce the relation model, a new surrogate model variant that we previously proposed~\cite{haoBinaryRelationLearning2020}. The relation model diverges from traditional regression and classification models in its learning objective: it doesn't target the quality of a single solution (as in regression-based models) or the category of a solution (as in classification-based models), but rather the superiority relationship between two solutions. The relation model exploits the comparative nature of evolutionary algorithms~\cite{loshchilov2010comparison} and has demonstrated remarkable performance in single-objective~\cite{haoBinaryRelationLearning2020} and multi-objective problems~\cite{hao2022expensive,hao2021approximated,hao2023relation,yuan2021expensive,tian2023pairwise}.

In this study, we strive to customize the construction strategy of the relation model to fulfill the framework's demand for model selection accuracy amidst the requirement for potential quality solutions. Therefore, we propose a dual relation models-assisted single-objective optimization algorithm (DRSO) and design two methods for constructing relation models. These methods respectively select the optimal solution ($\mathcal{Q}_{best}$) and high-quality unevaluated solutions~($\mathcal{P}_u$). We employ the distribution estimation algorithm (EDA) to study the population's distribution information and generate offspring solutions. While the strategy of utilizing unevaluated solutions has been implemented for multi-objective optimization~\cite{zhang2022dual}, our current work specifically focuses on designing a relation model for the selection of unevaluated solutions in single-objective optimization, instead of using a classifier. The main contributions of this paper can be summarized as follows:

\begin{itemize}
  \item Illumination of the issue of offspring quality degradation in SAEAs when only a single offspring per generation is selected. In response, we propose a simple and universal method fueled by unevaluated solutions.
  \item Proposal of two methods for constructing relation models, known as the fitness-based and category-based criteria. These methods leverage data relationships to construct surrogate models.
  \item Introduction of a novel strategy, based on the EDA, for generating solutions by integrating evaluated and unevaluated solutions. The efficacy of this novel algorithm is validated on two test suites, highlighting both the effectiveness of the relation model and the significance of incorporating unevaluated solutions.
  \end{itemize}

The rest of the article unfolds as follows. Section~\ref{sec:preliminaries} presents some preliminaries. Section~\ref{sec:proposedalg} outlines the unevaluated solutions driven SAEAs framework, covering the construction of the relation model and the generation of trial solutions. Section~\ref{sec:experiment} showcases an empirical evaluation of the proposed method and compares it with other methods across two test suites. Finally, Section~\ref{sec:conclusions} provides a summary of the paper and explores potential directions for future research.

\section{Preliminaries}
\label{sec:preliminaries}

This section provides the preliminary knowledge related to this work. Section~\ref{sec:eop} presents the basic concepts of EOPs. Section~\ref{sec:offspring_selection} introduces different strategies for offspring selection. Section~\ref{sec:surrogate} provides an overview of surrogate models, particularly focusing on relation models. Section~\ref{sec:population} discusses the impact of population variance on the efficiency of SAEAs.

\subsection{Expensive optimization problems}
\label{sec:eop}
An unconstrained minimization expensive optimization problem can be formulated as follow:
\begin{align}
 \label{equ:func}
 \min_{\mathbf{x} \in \Omega}~f(\mathbf{x}) 
 \end{align}
where $\mathbf{x}=(x_1,\ldots,x_n)^T$ is a decision variable vector, $\Omega \in R^n$ defined the feasible region of the search space. Given that $f: R^n \rightarrow R$ is the objective function, which is essentially a black-box due to the difficulty in tracking its internal workings, optimization problems in real-world applications that involve $f(\cdot)$ can be quite costly. In fact, the lack of a closed-form objective function and the expensive nature of evaluating $f(\cdot)$ pose significant challenges to both numerical and heuristic optimization techniques that are traditionally employed.

\subsection{Offspring Selection Methods}
\label{sec:offspring_selection}
The purpose of offspring selection is to guide population movement toward the optimal regions while ensuring a certain level of diversity in the distribution. Depending on the offspring selection strategy, representative works in SAEAs can be categorized into three groups as follows:

\begin{itemize}
\item select $N$: This strategy is employed during the iterative process of several algorithms such as BCPS~\cite{zhang2018preselection}, FCPS~\cite{zhou2019fuzzy}, RCPS~\cite{haoBinaryRelationLearning2020} and SAMFEO~\cite{LiExpensiveOptimizationSurrogateAssisted2022a}. With the use of the reproduction operator, it generates a significant number of trial solutions surpassing the population size ($N$). Following this, a surrogate model is applied to select $N$ solutions for real evaluation, creating the offspring solutions.
\item select $1$: In every generation, only the top solution is chosen for real function evaluation and preserved in an archive. Acquisition functions~\cite{} are employed to enhance the exploratory capability of the algorithm. Specifically, GPEME~\cite{journals/tec/LiuZG14} and SADE-Sammon~\cite{chen2020surrogate} utilize the lower confidence bound~(LCB)~\cite{srinivas2009gaussian} to guide the search, EGO~\cite{jones1998efficient} adopts the expected improvement~(EI) method~\cite{mockus1998application}, and SA-EDA~\cite{hao2022surrogate} integrates multiple acquisition strategies using the GP-Hedge method~\cite{DBLP:conf/uai/HoffmanBF11} to enhance the robustness of the selection.
\item others: Customized approaches have been proposed in SAMSO~\cite{journals/tcyb/LiCGS21} and SACOSO~\cite{sun2017surrogate}, where multi-particle swarm is utilized to increase diversity through mutual interactions between swarm. In LLSO~\cite{wei2020classifier} and DFC-MOEA~\cite{zhang2022dual}, a hierarchical strategy is employed for solution selection. In addition, LLSO enhances population diversity by introducing random solutions, while DFC-MOEA selects solutions with medium membership degrees using a classifier.
\end{itemize}

Each of the aforementioned methods has its own advantages, with the core consideration being the balance of interests under a limited computation budget of EOPs.

\subsection{Surrogate model}
\label{sec:surrogate}

In SAEAs, surrogate models typically fall into two main categories~\cite{haoBinaryRelationLearning2020}: regression and classification models. In regression-based SAEAs, the original function is replaced with a curve fitting the data points' distribution. Examples of such models include polynomial regression~\cite{lian2005multiobjective}, radial basis function~(RBF)~networks~\cite{journals/tec/SunJCDZ17}, and Gaussian processes~(GPs)~\cite{willmes2003comparing}. Classification-based SAEAs, on the other hand, label solutions based on their quality, using models such as support vector machines (SVM)~\cite{li2019ibea}, artificial neural networks (ANN)~\cite{journals/tec/PanHTWZJ19}, and fuzzy K-nearest neighbor (KNN)~\cite{conf/aaai/ZhouZSZ19}.

A newer category in SAEAs is relation learning~\cite{hao2022expensive,yuan2021expensive,haoBinaryRelationLearning2020,hao2021approximated,hao2023relation}, where the model is trained on the relationships between solutions, as opposed to using a single solution in regression or classification-based SAEAs. This approach shows promise in single-objective optimization, as it leverages the superiority and inferiority relationships between solutions for pre-selection operations on offspring solutions, resulting in improved performance~\cite{haoBinaryRelationLearning2020}. In multi-objective optimization, methods like REMO~\cite{hao2022expensive} and CREMO~\cite{hao2023relation} use a penalty-based boundary intersection (PBI)~\cite{zhang2007moea} approach to categorize solutions in the multi-objective space. A relation dataset is constructed based on the belonging relationship between samples, and a neural network is trained to learn the sample features. This process has proven effective in creating reliable surrogates for both continuous and discrete optimization problems. Methods like $\theta$-DEA-DP~\cite{yuan2021expensive} directly apply the dominance relationship as the relationship definition for solutions, focusing on the dominance relationship learning and prediction.

Previous studies have demonstrated the advantages of using relation models in SAEAs. The construction of the relation model can generally be divided into three steps: data preparation, model training, and model usage.
In data preparation, a certain criterion is used to construct relationship samples, $\mathcal{D} = {(\langle \mathbf{x}_{i},\mathbf{x}_{j} \rangle,l)| \mathbf{x}_{i}, \mathbf{x}_{j} \in \mathcal{P} }$, where $\langle \mathbf{x}_{i},\mathbf{x}_{j} \rangle$ is a feature vector composed of each pair of solutions, and $l$ is the label of $\langle \mathbf{x}_{i},\mathbf{x}_{j} \rangle$. Machine learning methods are then used to learn from the relation data, and a predict method based on the relation model is designed to select solutions. In this work, we address the specific needs of selecting the best solution~($\mathcal{Q}_{best}$) and high-quality unevaluated solutions~($\mathcal{P}_{u}$) and propose new methods for constructing relationship models.

\subsection{Impact of variance among adjacent generations' populations}
\label{subsec:variance}
\begin{figure}[htbp]
  \centering
  \subfloat{\includegraphics[width=1\textwidth]{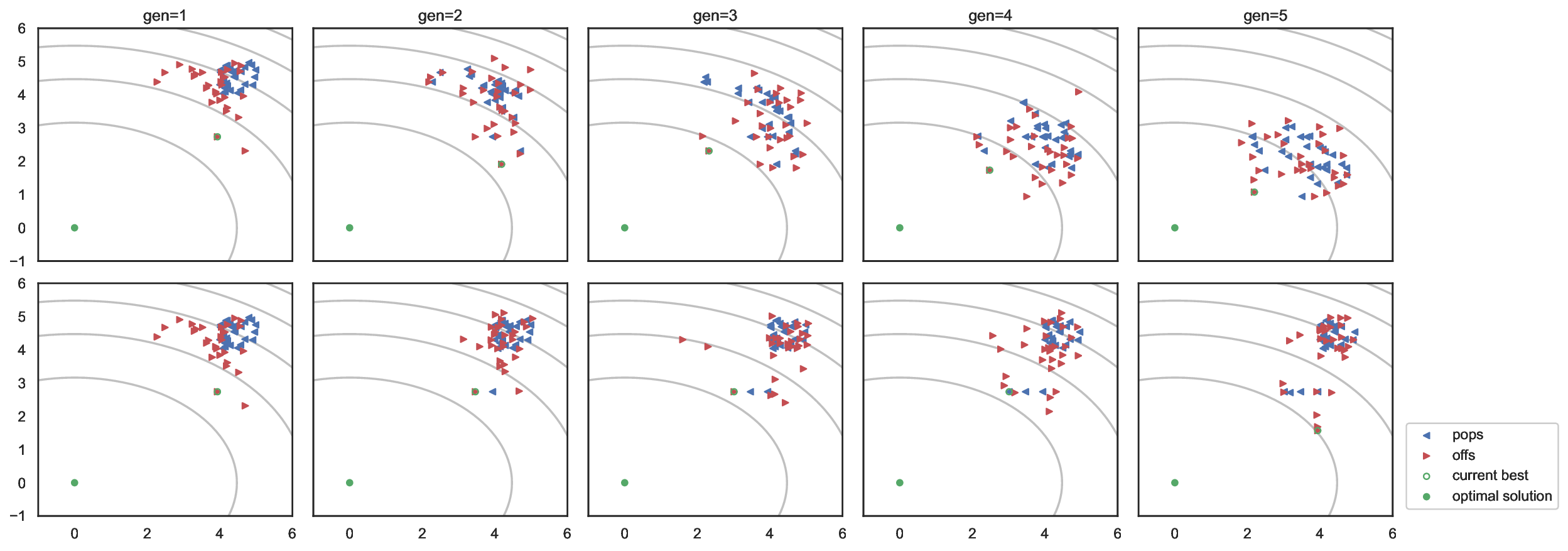}} \\
\caption{Distribution of the population during continuous evolution.}
\label{fig:ga_run_2d}
\end{figure}

When confronted with the EOP mentioned in Eq.~(\ref{equ:func}), it is expedient to solely evaluate the optimum solution in order to conserve the evaluation budget. However, this paradigm precipitates a decline in inter-population variance, thereby engendering new solutions in the subsequent iteration that are constrained within the present search region, culminating in a low-effectiveness predicament in the algorithm. 
Figure~\ref{fig:ga_run_2d} presents a visual representation of the results obtained from five successive generations of search on a 2-dimensional Ellipsoid function~\cite{journals/tec/LiuZG14} using the genetic algorithm~(GA)~\cite{holland1992adaptation}. The first row shows the selection of $N$ solutions per generation, whereas the second row illustrates the selection of only the optimal solution for the next generation. The outcomes indicate that utilizing a single solution to update the population can lower the search efficiency of the original GA algorithm. This is due to the fact that selecting only the best solution can result in a loss of diversity in the population and hinder the exploration of the search space. 

\label{sec:population}

Additionally, we carried out 30 independent runs utilizing GA, differential evolutionary~(DE)~\cite{storn1997differential}, and EDA~\cite{journals/tec/ZhouSZ15} (three fundamental EAs) on LZG test suite~\cite{journals/tec/LiuZG14}. According to the experimental results shown in Table~\ref{tab:select_num} and analyzed using the Wilcoxon rank sum test~\cite{hollander2013nonparametric}, selecting a single solution leads to a decrease in all algorithms performance. It can be seen from this that the problem of performance degradation caused by selecting only the optimal solution is commonly present in various evolutionary algorithms.
\begin{table*}[ht!]
\renewcommand{\arraystretch}{1.1}
\centering
\scriptsize
\caption{Statistics of mean and standard deviation results obtained by GA, ED and EDA on LZG test suite with $n=20$.}
\begin{tabular}{cccccc}
\toprule
  Alg                 & method   & Ellipsoid              & Rosenbrock              & Ackley              & Griewank
  \\ \hline
  \multirow{2}{*}{GA} & select N & 3.97e-01(1.37e-01) & 5.17e+01(2.68e+01) & 2.25e+00(3.57e-01) & 1.01e+00(1.79e-03)          \\ 
                      & select 1 & 1.45e+01(7.57e+00)($-$) & 9.27e+01(4.01e+01)($-$) & 8.33e+00(2.84e+00)($-$) & 1.22e+00(1.02e-01)($-$)          \\ \hline
  \multirow{2}{*}{DE} & select N & 2.64e-01(8.02e-02) & 3.36e+01(1.51e+01) & 2.60e+00(4.07e-01) & 1.01e+00(2.39e-03)          \\
                      & select 1 & 3.27e+01(2.50e+01)($-$) & 1.29e+02(4.92e+01)($-$) & 9.42e+00(1.61e+00)($-$) & 1.57e+00(3.91e-01)($-$)          \\ \hline
  \multirow{2}{*}{EDA}  & select N & 4.62e-02(7.29e-02) & 1.92e+01(3.08e+00) & 3.69e-01(6.46e-01) & 9.61e-01(1.27e-02)          \\
                      & select 1 & 1.00e+01(2.82e+00)($-$) & 5.72e+01(1.18e+01)($-$) & 7.31e+00(5.81e-01)($-$) & 1.21e+00(4.18e-02)($-$)         \\ \hline             
\end{tabular}
\label{tab:select_num}
\end{table*}

The aforementioned toy studies demonstrate that a decrease in inter-population variance can lead to a decline in the performance of some fundamental algorithm operators. Therefore, adding some unevaluated solutions to supplement diversity can be a direct and simple method to improve the performance of the SAEAs algorithm.

\section{Proposed method}
\label{sec:proposedalg}
In this section, we begin by introducing a basic framework for surrogate-assisted selection and unevaluated solution driven reproduction. Then, we will present two innovative approaches for constructing relation models. Finally, we will provide a detailed explanation of the reproduction process within this framework.

\subsection{Main Framework}

\begin{algorithm}
  \renewcommand{\algorithmicrequire}{\textbf{Input:}}
  \renewcommand{\algorithmicensure}{\textbf{Ouput:}}
  \caption{Main Framework}
  \begin{algorithmic}[1]
    \REQUIRE $N$~(population size); $FE_{max}$~(maximum number of FEs); $\alpha$~(size of training data set).
    \ENSURE $\mathcal{A}_{best}$~(the optimum solution).
    \STATE $\mathcal{P}_e \leftarrow \boldsymbol{Initialization}(N)$. \hspace{6.54cm} \it{ /* initialize population*/} 
    \STATE$\mathcal{A} \leftarrow \mathcal{P}_e$.   \hspace{10.1cm} \it{ /* update archive */} 
    \STATE $fes \leftarrow N$. \hspace{8.3cm} \it{/* update evaluation counter */}
    \STATE $\mathcal{P}_u \leftarrow \emptyset$.  \hspace{9cm} \it{ /* initialize an empty set */} 
    \WHILE{$fes\leq FE_{max}$}   
    \STATE $Q\leftarrow \boldsymbol{Reproduction}(\mathcal{P}_e,\mathcal{P}_u,N)$. \hspace{4.7cm} \it{ /* generate new solutions */}
    \STATE $\mathcal{M} \leftarrow \boldsymbol{Training}(\mathcal{A}_{1:\alpha})$. \hspace{6.29cm} \it{ /* train surrogate model */}  \label{alg:framework_training}
    \STATE $[\mathcal{Q}_{best},\mathcal{P}_u] \leftarrow \boldsymbol{SA\_selection}(\mathcal{Q},\mathcal{M})$. \hspace{3.73cm} \it{ /* surrogat-assisted selection */} \label{alg:framework_sa}
    \STATE $\mathcal{A} \leftarrow \mathcal{A} \cup \boldsymbol{Evaluation}(\mathcal{Q}_{best})$. \hspace{2.31cm} \it{ /* evaluate new solution and update archive */}
    \STATE $\mathcal{P}_e \leftarrow \mathcal{A}_{1:N}$. \hspace{8.85cm} \it{ /* update population */}
    \STATE $fes \leftarrow fes+1$. \hspace{7.05cm} \it{ /* update evaluation counter */} 
    \ENDWHILE
  \end{algorithmic}
  \label{alg:framework}
\end{algorithm}

Algorithm~\ref{alg:framework} presents a basic framework put forward in this article, comprising surrogate-assisted selection and unevaluated solution driven reproduction. The specifics are succinctly summarized as follows.

\begin{itemize}
\item \textbf{Initialization~(lines 1-4)}: A set of $N$ initial solutions are sampled from $\Pi_{i=1}^n\left[a_i, b_i\right]$ by means of the Latin hypercube sampling method~(LHS)~\cite{mckay2000comparisona}, with each of these solutions undergoing an evaluation by the real function and subsequently being stored in the archive $\mathcal{A}$. The fitness evaluation count of these evaluations, denoted by the $fes$, is updated accordingly. Eventually, an empty set $\mathcal{P}_u$ needs to be initialized to store the unevaluated solutions selected by the surrogate model in the subsequent steps.
\item \textbf{Stop condition~(line 5)}: The algorithm halts once the $fes$ surpasses the designated maximum number of evaluations~($FE_{max}$).
\item \textbf{Generate new solutions~(line 6)}: Based on the current evaluated population $\mathcal{P}_e$ and unevaluated population $\mathcal{P}_u$, an offspring population $\mathcal{Q}$ containing $N$ individuals is generated utilizing various heuristic operators, such as DE, GA, EDA, among others. In this study, an approach combining a variable-width histogram~(VWH) model and local search will be employed to generate new solutions~\cite{journals/tec/ZhouSZ15}.
\item \textbf{Train surrogate model~(line 7)}: Selecting the optimal $\alpha$ solutions from archive $\mathcal{A}$, surrogate models will be trained. In this work, two customized methods for constructing relation models will be provided.
\item \textbf{Surrogate-assisted selection~(line 8)}: The surrogate model is utilized to evaluate the solutions in the offspring population $\mathcal{Q}$, with the optimal solution being selected as $\mathcal{Q}_{best}$. A portion of the high-quality solutions in $\mathcal{Q}$ will be selected as unevaluated solutions, restored in $\mathcal{P}_u$.
\item \textbf{Update archive~(line 9)}: The solution $\mathcal{Q}_{best}$ will be evaluated by the real objective function and saved in archive $\mathcal{A}$.
\item \textbf{Select solution for next generation~(line 10)}: $N$ evaluated solutions are selected from the archive $\mathcal{A}$ based on their objective function values to constitute the population $\mathcal{P}_e$.
\item \textbf{Update the counter~(line 11)}: Since only one solution, $\mathcal{Q}_{best}$, undergoes real evaluation during each iteration, $fes$ is incremented by one.
\end{itemize}

In order to facilitate the model-assisted selection (line~\ref{alg:framework_sa}), it is necessary to devise surrogate models that can accurately select the optimal solution $\mathcal{Q}_{best}$ from $\mathcal{Q}$, as well as identify a subset of potentially good solutions that have not been evaluated but meet a certain threshold to form $\mathcal{P}_u$. Additionally, we need to design a method to generate offspring solutions using these unevaluated solutions. Therefore, in the following sections, we will provide a detailed description of the design of the surrogate model and the generation of new solutions.

\subsection{Relation model}
This subsection proposes two relation-based methods for constructing surrogate models, which are referred to as the fitness-based criterion (C1) and the category-based criterion (C2), respectively, and are used for two specific applications. The C1 criterion is used for selecting $\mathcal{Q}_{best}$, while the C2 criterion is used for selecting $\mathcal{P}_u$. Each model consists of three components: data preparation, model training, and model usage. The following sections will provide a detailed description of the implementation details of each component.

\subsubsection{Data preparation}

Data preparation refers to how to construct relation pairs from the original training data $\mathcal{D}$. We have designed two data construction methods for C1 and C2 criteria.
\begin{algorithm}[ht!]
  \renewcommand{\algorithmicrequire}{\textbf{Input:}}
  \renewcommand{\algorithmicensure}{\textbf{Ouput:}}
  \caption{Data preparation in fitness criterion~(C1)}
  \label{alg:getR_C1}
  \begin{algorithmic}[1]
    \REQUIRE $\mathcal{D}= \{(\mathbf{x}_{1},f(\mathbf{x}_{1})),\cdots,(\mathbf{x}_{1},f(\mathbf{x}_{\alpha}))\}$~(\text{Training Data}).
    \ENSURE $\mathcal{D}_r = \{ (\langle\mathbf{x}_{i},\mathbf{x}_{j}\rangle,l)| i,j\in[1,\alpha],l \in [-1,+1] \}$~(\text{Relation Data}).
    \STATE $\mathcal{D}_r \leftarrow \{(\langle \mathbf{x}_{i},\mathbf{x}_{j} \rangle,l)| \mathbf{x}_i, \mathbf{x}_j \in P, i\neq j\}$, where the label $l$ is assigned as follow:
    $$l(\langle \mathbf{x}_{i},\mathbf{x}_{j}\rangle)=\begin{cases}
      +1, & f(\mathbf{x}_i) <f(\mathbf{x}_j) \\
      -1, & f(\mathbf{x}_j) \geqslant f(\mathbf{x}_i)\\
    \end{cases}$$ \\ 
  \end{algorithmic}
  \end{algorithm}

  \begin{algorithm}
  \renewcommand{\algorithmicrequire}{\textbf{Input:}}
  \renewcommand{\algorithmicensure}{\textbf{Ouput:}}
  \caption{Data preparation in category criterion~(C2)}
  \begin{algorithmic}[1]
    \REQUIRE $\mathcal{D}= \{(\mathbf{x}_{1},f(\mathbf{x}_{1})),\cdots,(\mathbf{x}_{1},f(\mathbf{x}_{\alpha}))\}$~(\text{Training Data});
    $t$~(Classification Threshold).
    \ENSURE $\mathcal{D}_r = \{ (\langle\mathbf{x}_{i},\mathbf{x}_{j}\rangle,l)| i,j\in[1,\alpha],l \in [-1,0,+1]\}$~(\text{Relation Data}).
  
    \STATE $X_{good}\leftarrow {x \in \mathcal{D}_{Top(t)}}$.
    \STATE $X_{bad}\leftarrow {x \in \mathcal{D}\land x\notin \mathcal{D}_{Top(t)}}$. 
    \STATE $\mathcal{D}_r \leftarrow \{(\langle \mathbf{x}_{i},\mathbf{x}_{j} \rangle,l)| \mathbf{x}_i, \mathbf{x}_j \in P, i\neq j\}$, where the label $l$ is assigned as follow:
    $$
    l(\langle \mathbf{x}_{i},\mathbf{x}_{j}\rangle)=\begin{cases}
    +1, & \mathbf{x}_i \in \mathbf{X}_{good}, \mathbf{x}_j \in \mathbf{X}_{bad}\\
    -1, & \mathbf{x}_i \in \mathbf{X}_{bad}, \mathbf{x}_j \in X_{good}\\
    +0,  & \mathbf{x}_i \in \mathbf{X}_{good}, \mathbf{x}_j \in \mathbf{X}_{good} \\
    -0, & \mathbf{x}_i \in \mathbf{X}_{bad}, \mathbf{x}_j \in \mathbf{X}_{bad}
    \end{cases}$$ 
    \STATE $\mathcal{D}_r \leftarrow\boldsymbol{LabelBalanced}(\mathcal{D}_r$).
  \end{algorithmic}
  \label{alg:getR_C2}
  \end{algorithm}

\begin{itemize}
\item Fitness-based criterion~(C1): To determine the superiority or inferiority between any given pairs of relations $\langle x_i,x_j\rangle$, their corresponding fitness values (i.e., objective function values) are used as a pivotal criterion. This allows for the assignment of a label to each pair. The process is elaborated in Algorithm~\ref{alg:getR_C1}, which generates a labeled training dataset $\mathcal{D}_{r}$ consisting of two classes. Here, $\alpha$ denotes the total number of elements present in the dataset $\mathcal{D}$.
\item Category-based criterion~(C2): First, a threshold is set based on the distribution of fitness values in the current population. Then, according to the comparison between the solution in $\mathcal{D}$ and the threshold, they are classified into different categories ($\mathbf{X}_{good}$ and $\mathbf{X}_{bad}$). Finally, labels are assigned to the relation pairs $\langle \mathbf{x}_i,\mathbf{x}_j \rangle$ based on the categories that make up the solution for the pairs. The specific details are shown in Algorithm~\ref{alg:getR_C2}. In lines 1-2, based on the classification threshold $t$, the $t$ top solutions in the data set $\mathcal{D}$ are selected as $\mathbf{X}_{good}$ samples according to their fitness values from best to worst. while the rest are assigned as $\mathbf{X}_{bad}$ samples. In line 3, the relation pair $\langle \mathbf{x}_{i}, \mathbf{x}_{j}\rangle $ is assigned a label according to the categories to which they belong. Since $t$ is not necessarily equal to $50\%$, the labels of the pairs in dataset $\mathcal{D}_r$ may not be balanced. To address this, we have further implemented the balancing strategy~(line 4) proposed in \cite{hao2022expensive}.
\end{itemize}

The label balancing strategy is described as follows. Let $L(+1)$, $L(-1)$, $L(+0)$, and $L(-0)$ represent the sets of pairs labeled as `+1', `-1', `+0' and `-0' respectively. The symbol $| \cdot \vert$ denotes the cardinality of a set. It is apparent that $|L(+1)| = |L(-1)|$, and $(|L(+0)| + |L(-0)|) > |L(+1)|$. In order to balance the training data sets, certain points from $L(+0)\cup L(-0)$ must be removed. Let $|L(+1)| = |L(-1)| = \theta$. There exist three situations.

\begin{itemize}
  \item If $|L(+0)| > 0.5\theta$ and $|L(-0)| > 0.5\theta$, $0.5\theta$ points are arbitrarily retained from both $L(+0)$ and $L(-0)$.
  \item If $|L(+0)| > 0.5\theta$ and $|L(-0)| < 0.5\theta$, $L(-0)$ is retained, and $\theta - |L(0)|$ points are randomly selected from $L(+0)$.
  \item If $|L(+0)| < 0.5\theta$ and $|L(-0)| > 0.5\theta$, $L(+0)$ is retained, and $\theta - |L(+0)|$ points are randomly selected from $L(0)$.
  By following this method, the three training data sets all have a size of $\theta$.
\end{itemize}

After employing two data preparation strategies and customizing the training data based on the C1 and C2 criteria, we have generated a 2-class dataset for $\mathcal{D}_r$ using the C1 strategy and a 3-class dataset using the C2 strategy. In the following section, we will introduce the model training process.

\subsubsection{Model training}

Extreme Gradient Boosting~(XGBoost)~\cite{chen2016xgboost} is a machine learning algorithm that is widely used in various data-driven applications. XGBoost is based on the concept of gradient boosting, where weak models are combined to create a strong model. In this work, XGBoost was used to learn the data features of $\mathcal{D}_r$.

The relation pair samples $\langle \mathbf{x}_i,\mathbf{x}_j \rangle$ are the data features of $\mathcal{D}_r$, and the label $l$ indicates the relationship between two solutions in a set of pairs. The model $\mathcal{M}$ is trained using the XGBoost algorithm, as shown in Eq.~(\ref{equ_RESO.model}).
\begin{equation}
  \label{equ_RESO.model}
l = \mathcal{M}(\langle \mathbf{x}_i,\mathbf{x}_j \rangle)
\end{equation}
In line~\ref{alg:framework_training} of Algorithm~\ref{alg:framework}, two models need to be trained for the two criteria, hence we differentiate them as $\mathcal{M}_1$ and $\mathcal{M}_2$. The next step is to explain how to select the potential solutions based on the two models.

\subsubsection{Molde usage}

For selecting appropriate solutions based on the two models $\mathcal{M}_1$ and $\mathcal{M}_2$, we propose two different model usage strategies, corresponding to the selection of the $\mathcal{Q}_{best}$ and $\mathcal{P}_u$. Specifically, we adopt the basic idea of `voting-scoring' used in previous works~\cite{hao2022expensive} and redesign the rules for its implementation.

The term `vote' pertains to the prediction process of labeling $\langle \mathbf{u},\mathbf{x} \rangle$ and $\langle \mathbf{x},\mathbf{u} \rangle$, where an unknown solution $\mathbf{u}$ and an evaluated solution $\mathbf{x}$ are combined. This procedure can be regarded as an assessment of the unknown solution's quality based on the quality of the known solution $\mathbf{x}$. As such, we refer to this process as a `voting' mechanism. The `score' is determined based on the voting outcomes of all solutions $\mathbf{x}$ in the training dataset $\mathcal{D}$, and a specific rule is employed for statistical analysis. The rule's configuration necessitates consideration of the position between $\mathbf{x}$ and $\mathbf{u}$, as well as $\mathbf{x}$'s fitness or category. Next, we will introduce the `vote-score' strategies that are devised based on the C1 and C2 criteria.

\begin{itemize}
\item Fitness-based criterion~(C1): For a newly generated solution $\mathbf{u}\in\mathcal{Q}$, it combines all the evaluated solutions $\mathbf{x} \in \mathcal{D}$. Based on the positions of the two solutions, two sets of relation pairs can be obtained, e.g., $\langle \mathbf{x},\mathbf{u} \rangle$ and $\langle \mathbf{u},\mathbf{x} \rangle$. Thus, utilizing Eq.~(\ref{equ:C1_voting}), two sets of predicted outcomes, $l^{\RNum{1}}$ and $l^{\RNum{2}}$, can be derived. 

\begin{equation}
\label{equ:C1_voting}
\begin{split}
l^{\RNum{1}}=&~\{\mathcal{M}_1(\langle \mathbf{x},\mathbf{u}\rangle),\mathbf{x}\in \mathbf{X}\} \\
l^{\RNum{2}}=&~\{\mathcal{M}_1(\langle \mathbf{u},\mathbf{x}\rangle),\mathbf{x}\in \mathbf{X}\} \\
\end{split}
\end{equation}
The scoring rules are defined by Eq.~(\ref{equ:C1_scoring}).
\begin{equation}
\label{equ:C1_scoring}
\begin{split}
S_1(\mathbf{u}) =c(l^{\RNum{2}}_{+1})+c(l^{\RNum{1}}_{-1})-c(l^{\RNum{1}}_{+1})-c(l^{\RNum{2}}_{-1})
\end{split}
\end{equation}
Here, the function $c(\cdot)$ returns the cardinality of elements present in the input set. The subscript of $l$ denotes the labels of the relation pairs that constitute the current subset. For example, $l^{\RNum{1}}_{+1}$ denotes a set that encompasses all elements in the set $l^{\RNum{1}}$ whose predicted label equals $+1$. The quality of solution $\mathbf{u}$ can be assessed by utilizing Eq.~(\ref{equ:C1_scoring}), where a higher value indicates superior quality of $\mathbf{u}$. Under the C1 criterion, the ultimate learning outcome can be perceived as a regression process for the original data distribution.

\item Category-based criterion~(C2): Under the C2 criterion, the `voting' rule is formulated as Eq.~(\ref{equ:C2_voting}). As $\mathbf{x}$ possesses a categorical attribute (`good', `bad'), the voting outcomes are classified into four categories based on the position and category of $\mathbf{x}$. The relation model $\mathcal{M}_2$ forecasts the outcomes of the four groups of relation pairs, denoted by set $l^{\RNum{1}}$, $l^{\RNum{2}}$, $l^{\RNum{3}}$, and $l^{\RNum{4}}$, respectively.

\begin{equation}
  \label{equ:C2_voting}
  \begin{split}
  l^{\RNum{1}}=&~\{\mathcal{M}_2(\langle \mathbf{x},\mathbf{u}\rangle),\mathbf{x}\in \mathbf{X}_{good}\} \\
  l^{\RNum{2}}=&~\{\mathcal{M}_2(\langle \mathbf{u},\mathbf{x}\rangle),\mathbf{x}\in \mathbf{X}_{good}\} \\
  l^{\RNum{3}}=&~\{\mathcal{M}_2(\langle \mathbf{x},\mathbf{u}\rangle),\mathbf{x}\in \mathbf{X}_{bad}\} \\
  l^{\RNum{4}}=&~\{\mathcal{M}_2(\langle \mathbf{u},\mathbf{x}\rangle),\mathbf{x}\in \mathbf{X}_{bad}\} \\
  \end{split}
  \end{equation}
  The scoring rules are defined by Eq.(\ref{equ:C2_scoring}).
  \begin{equation}
    \label{equ:C2_scoring}
    \begin{split}
    S_{2}(\mathbf{u}) = \frac{1}{|\mathbf{X}|}\times(c(l^{\RNum{2}}_{+1})+c(l^{\RNum{4}}_{+1})+c(l^{\RNum{1}}_{0})+c(l^{\RNum{2}}_{0})+c(l^{\RNum{1}}_{-1})+c(l^{\RNum{3}}_{-1})
    \\-c(l^{\RNum{1}}_{+1})-c(l^{\RNum{3}}_{+1})-c(l^{\RNum{3}}_{0})-c(l^{\RNum{4}}_{0})-c(l^{\RNum{2}}_{-1})-c(l^{\RNum{4}}_{-1}))
    \end{split}
    \end{equation}
\end{itemize}
In Eq.~(\ref{equ:C2_scoring}), the symbolism is similar to that of Eq.~(\ref{equ:C1_scoring}), but with a focus on the processing of the `0' label. According to the definition of relation pairs in the C2 criterion~(Algorithm~\ref{alg:getR_C2}), the `0' label indicates that the two solutions in the pair belong to the same category. Therefore, based on the category of $\mathbf{x}$, the contribution to the scoring can be determined. For instance, $l^{\RNum{2}}_{+1}$ denotes the prediction result of $\langle \mathbf{u},\mathbf{x} \rangle$ as `+1', indicating that $\mathbf{u}$ is considered better than $\mathbf{x}$. As a result, the score $c(l^{\RNum{2}}_{+1})$ has a positive impact on the quality of $S_2(\cdot)$. $S_2(\cdot)$ can be scaled to $[-1, +1]$ by multiplying it with $\frac{1}{|\mathbf{X}|}$. When $S_2(\mathbf{u}) > 0$, it indicates that the relation model considers the current solution $\mathbf{u}$ to be in the `good' category, whereas when $S_2(\mathbf{u}) < 0$, it indicates that the relation model considers the current solution $\mathbf{u}$ to be in the `bad' category. Moreover, the larger the $|S_2(\mathbf{u})|$ value, the greater the likelihood of belonging to either of the two categories. Under the C2 criterion, the final learning outcome can be viewed as a classification process for the original data distribution.

After processing the features of the original training data (data preparation) and training the models, we obtain two models $\mathcal{M}_1$ and $\mathcal{M}_2$ based on the relation of the solutions. These models can be used to select solutions in line~\ref{alg:framework_sa} of Algorithm~\ref{alg:framework}. Specifically, each solution in the offspring population $\mathcal{Q}$ will be predicted by $\mathcal{M}_1$ and $\mathcal{M}_2$, and then based on the C1 criterion, the solution with the maximum $S_1$ value will be selected as $\mathcal{Q}_{best}$, and based on the C2 criterion, all solutions that satisfy $S_2>0$ will be selected as the $\mathcal{P}_u$ population.

\subsection{Reproduction}

This work employs the EDA/LS, proposed by Zhou et al~\cite{journals/tec/ZhouSZ15}, as the fundamental method for generating new solutions, while incorporating information from the unevaluated solutions in population $\mathcal{P}_u$ to generate offspring population $\mathcal{Q}$. The EDA/LS algorithm includes two key operators, namely the variable-width histogram~(VWH) and the local search method that combines global statistical information with individual location information to improve the performance of an EDA. First, a brief introduction to the VWH is presented, followed by an explanation of the local search method. Finally, the method for integrating unevaluated solutions to generate offspring population $\mathcal{Q}$ is described.

\subsubsection{Variable-width histogram model}

An Estimation of Distribution Algorithm (EDA) is an evolutionary algorithm variant that uses a probabilistic model to guide the search toward promising regions of the search space. It replaces traditional crossover or mutation operators with this model. A specific type, Variable-width histograms (VWH)~\cite{journals/tec/ZhouSZ15}, assumes no cross-dimensional correlations and uses a histogram model to track the population distribution. VWH emphasizes promising areas, reducing probabilities in other regions to prevent premature convergence, making it ideal for enhancing convergence in EOPs.

\begin{figure}[htbp]
  \centering
  \subfloat[early stage]{\label{fig:eda1} \includegraphics[width=0.3\textwidth]{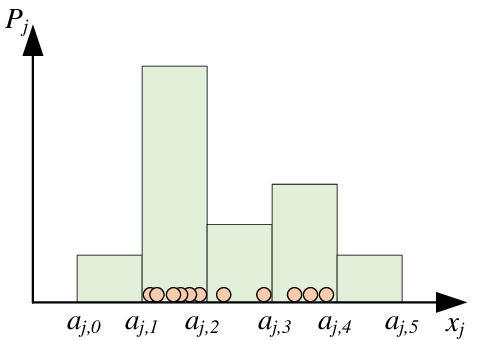}}
  \subfloat[late stage]{\label{fig:eda2}\includegraphics[width=0.3\textwidth]{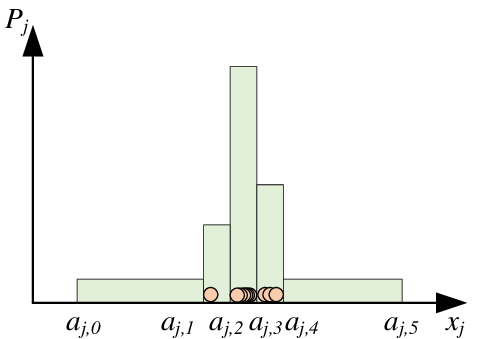}}
\caption{Illustration of VWH model for population on early and late search stage.}
\label{fig:EDA_MOEDA}
\end{figure}

Fig.~\ref{fig:EDA_MOEDA} illustrates the process of VWH. For the $j$-th variable, the search space $[a_j,b_j]$ is partitioned into $M$ bins, where the M-2 bins in the middle correspond to the regions with solutions in the current population $\mathcal{P}$. The values of the bins are determined by the number of solutions in each bin's interval, while the first and the last bins are assigned a lower value. To generate a new solution, a bin is randomly selected based on its value, and then a uniform random value is sampled from the selected bin's interval as the value of the new solution for the j-th variable. This process is repeated $n$ times to obtain a complete solution in the probability model VWH. By repeating this process $N$ times, $N$ offspring solutions are generated. For details on the modeling and sampling process, please refer to ~\cite{journals/tec/ZhouSZ15}. It is worth noting that the modeling and sampling stages of VWH only use the distribution information in the decision space, making it suitable to incorporate unevaluated solutions to update VWH.

\subsubsection{Local search}

In order to compensate for the lack of local solution information, EDA/LS~\cite{journals/tec/ZhouSZ15} proposes incorporating the results of local search into the offspring generated by the VWH model, given that the EDA model only uses the global information of the population to generate new solutions. In particular, a local model is constructed based on some of the best solutions from the current population $\mathcal{P}$, which is then utilized to generate a portion of the new solutions. Afterward, these solutions are randomly combined with the solutions sampled from VWH to form the final offspring population $\mathcal{Q}$. For more details, please refer to EDA/LS. Only the evaluated solutions are used for local search in this work, as they are driven by objective values.

\subsubsection{Unevaluated solutions driven reproduction}

\begin{figure}[ht!]
  \centering
  \subfloat{\includegraphics[width=.65\textwidth]{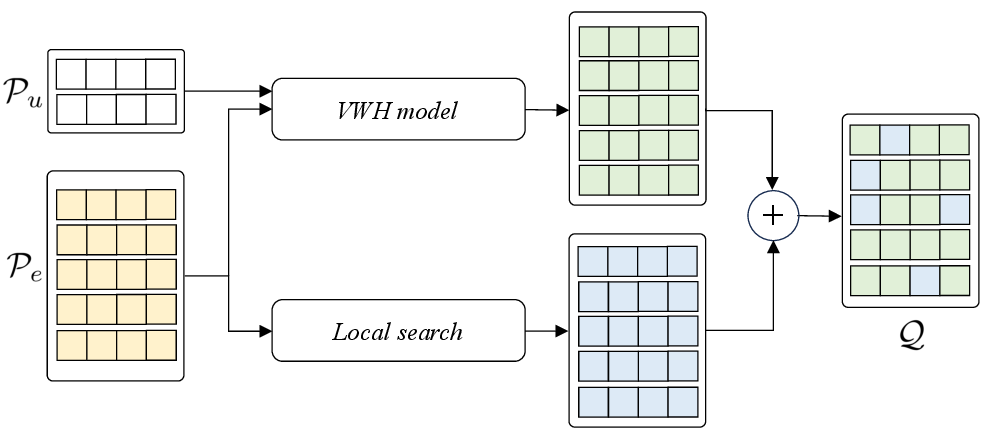}} \\
\caption{Flowchart for generating new solutions.}
\label{fig:reproduction}
\end{figure}

In each iteration, the process of generating the offspring population, using a combination of VWH and local search with both $\mathcal{P}_e$ and $\mathcal{P}_u$, will be executed according to the flowchart illustrated in Fig~\ref{fig:reproduction}.

The two populations, one consisting of evaluated solutions and the other consisting of unevaluated solutions, will be merged and modeled using the VWH model to capture their distribution. The resulting distribution will be sampled to generate a new population. Since the VWH only utilizes information about the search space, whether a solution has been evaluated or not does not affect the operation of the model. The local search method only uses the population $\mathcal{P}_e$ to generate a new population, which is then randomly merged with the new population generated by the VWH model to obtain the final offspring population $\mathcal{Q}$. The implementation details and parameter settings of the VWH model, as well as the local search method and the ratio for merging the two temporary populations, will be set to the default values specified in EDA/LS~\cite{journals/tec/ZhouSZ15}.

\section{Experimental study}
\label{sec:experiment}

In this section, we will evaluate the performance of the proposed algorithm and the relation model through comprehensive numerical experiments. Specifically, these experiments encompass comparative studies, ablation studies, and further analyses of the relation model and unevaluated solutions.

\subsection{Experimental Settings}
\label{sec:settings}

\subsubsection{Test suites}

In the empirical study, we utilized two well-known test suites. The first test suite, LZG~\cite{journals/tec/LiuZG14}, consists of four test functions: Ellipsoid, Rosenbrock, Ackley, and Griewank. These functions exhibit a range of landscapes, including unimodal, gully, and multi-modal. The second test suite used was the YLL test suite~\cite{yao1999evolutionary}, which contains functions F1 through F4 with unimodal landscapes, and F5, F8 through F13 with multimodal landscapes. Function F6 has a step landscape, while function F7 has random noise. We evaluated the problems in both test suites in dimensions $n=20$ for small-scale and $n=50$ for median-scale.

\subsubsection{Algorithms in study}

For the empirical study, seven algorithms have been selected, namely CMA-ES~\cite{hansen2001completely}, FCPS-CoDE~\cite{conf/aaai/ZhouZSZ19}, EDA/LS~\cite{journals/tec/ZhouSZ15}, SAMSO~\cite{journals/tcyb/LiCGS21}, Skopt~\footnote{https://github.com/scikit-optimize/scikit-optimize}, GPEME~\cite{journals/tec/LiuZG14}, and DRSO. These algorithms can be classified into three categories.

\begin{itemize}
  \item Basic EAs: CMA-ES and EDA/LS are two generic EAs, not explicitly tailored for expensive optimization.
  \item Bayesian optimization: Skopt is an effective global optimization algorithm that operates within the Bayesian optimization framework. It employs GPs as the surrogate model.
  \item Surrogate-assisted EAs: FCPS-CoDE utilizes a fuzzy K-nearest neighbor-based classification model for evaluating candidate solutions. GPEME employs GPs for the evaluation of candidate solutions. SAMSO is a surrogate-assisted PSO algorithm that incorporates RBFs. DRSO is a dual relation models-assisted EDA that incorporates unevaluated solutions to generate new candidate solutions which are proposed in this work.
\end{itemize}

Due to the high computational complexity of Gaussian processes in high-dimensional spaces, GPEME and Skopt were only compared in the experiments for $n=20$.

\subsubsection{Parameters setting}

To ensure a fair comparison in the empirical study, we employ the recommended parameters specified in the original literature for each algorithm~\footnote{CMA-ES and SAMSO are implemented in Platemo~\cite{tian2017platemo}; Skopt:https://github.com/scikit-optimize/scikit-optimize; FCPS-CoDE and GPEME are implemented by us based on the original report.}. The specifics of these parameters are outlined below.

\begin{itemize}
  \item Termination condition: The maximum number of function evaluations ($FE_{max}$) is employed as the termination condition, set at 500 for all instances.
  \item Population size: Set $N=30$ for CMA-ES, EDA/LS, and FCPS-CoDE, Set $N=40$ for SAMSO~(default set in PlatEMO~\cite{tian2017platemo}). Set $N=50$ for GPEME and DRSO.
  \item DRSO employs $t=50\%$ for the C2 criterion to choose $\mathcal{P}_u$.
  \item Parameters of compared algorithms: default setting according to the original version.
\end{itemize}

Each algorithm is executed on each test instance for 30 independent runs to account for randomness. The initial step involves evaluating the independence of the results generated by the algorithms for each test instance using the Friedman test~\cite{friedman1937use}. The Wilcoxon rank sum test~\cite{hollander2013nonparametric} is employed to compare the results. In the tables, the symbols `$+$', `$-$', `$\thicksim$' signify that the value achieved by an algorithm is smaller than, greater than, or similar to the value obtained by the DRSO, at a significance level of 0.05.

  \begin{table*}[ht!]
    \renewcommand{\arraystretch}{1.1}
    \renewcommand{\tabcolsep}{2pt}
    \centering
    \caption{Statistics of mean and standard deviation results obtained by seven comparison algorithms on LZG and YLL test suites with $n=20$.} \scriptsize
    \begin{tabular}{ccccccccc}
    \toprule
    problem & p-value & CMA-ES & FCPS-CoDE & EDA/LS & Skopt & GPEME & SAMSO & DRSO \\
    \midrule
    \multirow{2}{*}{Ellipsoid} & \multirow{2}{*}{1.80e-33} & 1.50e+02[7]($-$) & 1.30e+02[6]($-$) & 7.17e+01[5]($-$) & \hl{5.52e-02[1]($+$)} & 3.20e-01[2]($+$) & 1.87e+01[4]($\approx$) & 6.17e+00[3] \\ 
     &  & (4.25e+01) & (3.13e+01) & (1.56e+01) & (1.91e-02) & (1.77e-01) & (2.52e+01) & (4.57e+00) \\  \hline
    \multirow{2}{*}{Rosenbrock} & \multirow{2}{*}{4.94e-32} & 3.03e+02[6]($-$) & 3.22e+02[7]($-$) & 2.37e+02[5]($-$) & 5.22e+01[2]($+$) & 1.27e+02[4]($-$) & \hl{3.57e+01[1]($+$)} & 1.02e+02[3] \\ 
     &  & (7.50e+01) & (1.05e+02) & (4.02e+01) & (1.21e+01) & (4.32e+01) & (2.47e+01) & (2.94e+01) \\  \hline
    \multirow{2}{*}{Ackley} & \multirow{2}{*}{2.45e-34} & 1.59e+01[6]($-$) & 1.48e+01[5]($-$) & 1.33e+01[4]($-$) & 7.16e+00[3]($-$) & \hl{3.77e+00[1]($+$)} & 1.83e+01[7]($-$) & 6.08e+00[2] \\ 
     &  & (1.28e+00) & (1.00e+00) & (7.37e-01) & (3.27e-01) & (5.52e-01) & (1.33e+00) & (1.05e+00) \\  \hline
    \multirow{2}{*}{Griewank} & \multirow{2}{*}{6.07e-34} & 5.09e+01[6]($-$) & 5.46e+01[7]($-$) & 2.96e+01[5]($-$) & \hl{1.02e+00[1]($+$)} & 1.17e+00[2]($+$) & 2.06e+01[4]($-$) & 3.08e+00[3] \\ 
     &  & (1.06e+01) & (1.33e+01) & (7.62e+00) & (1.52e-02) & (8.76e-02) & (1.33e+01) & (1.54e+00) \\  \hline
    \multirow{2}{*}{YLLF01} & \multirow{2}{*}{1.12e-34} & 6.32e+03[7]($-$) & 5.37e+03[6]($-$) & 3.17e+03[5]($-$) & \hl{2.07e+00[1]($+$)} & 1.67e+01[2]($+$) & 6.73e+02[4]($-$) & 2.47e+02[3] \\ 
     &  & (1.98e+03) & (1.86e+03) & (5.92e+02) & (1.36e+00) & (1.09e+01) & (6.43e+02) & (1.37e+02) \\  \hline
    \multirow{2}{*}{YLLF02} & \multirow{2}{*}{6.34e-26} & 1.15e+02[6]($-$) & 2.41e+01[3]($-$) & 2.67e+01[4]($-$) & \hl{-4.30e+18[1]($\approx$)} & 5.03e+02[7]($-$) & 3.27e+01[5]($-$) & 4.06e+00[2] \\ 
     &  & (2.20e+02) & (3.67e+00) & (3.93e+00) & (3.55e+18) & (1.04e+03) & (1.72e+01) & (1.60e+00) \\  \hline
    \multirow{2}{*}{YLLF03} & \multirow{2}{*}{9.88e-23} & 2.09e+04[5]($-$) & 1.26e+04[2]($+$) & 2.32e+04[6]($-$) & \hl{5.79e+03[1]($+$)} & 2.99e+04[7]($-$) & 1.71e+04[4]($\approx$) & 1.53e+04[3] \\ 
     &  & (5.68e+03) & (3.48e+03) & (4.21e+03) & (2.57e+03) & (6.03e+03) & (1.38e+04) & (3.94e+03) \\  \hline
    \multirow{2}{*}{YLLF04} & \multirow{2}{*}{5.24e-23} & 4.03e+01[5]($-$) & 4.17e+01[6]($-$) & 3.28e+01[4]($-$) & \hl{2.58e+01[1]($\approx$)} & 3.09e+01[3]($-$) & 7.06e+01[7]($-$) & 2.62e+01[2] \\ 
     &  & (1.25e+01) & (6.55e+00) & (3.26e+00) & (1.12e+01) & (8.94e+00) & (7.40e+00) & (7.99e+00) \\  \hline
    \multirow{2}{*}{YLLF05} & \multirow{2}{*}{4.69e-31} & 3.11e+06[6]($-$) & 4.71e+06[7]($-$) & 1.13e+06[5]($-$) & 1.65e+05[3]($-$) & 3.05e+05[4]($-$) & 1.12e+05[2]($-$) & \hl{6.52e+04[1]} \\ 
     &  & (1.43e+06) & (2.84e+06) & (5.76e+05) & (7.14e+04) & (1.60e+05) & (8.27e+04) & (8.91e+04) \\  \hline
    \multirow{2}{*}{YLLF06} & \multirow{2}{*}{1.00e-34} & 5.79e+03[6]($-$) & 6.26e+03[7]($-$) & 3.20e+03[5]($-$) & \hl{2.17e+00[1]($+$)} & 2.32e+01[2]($+$) & 8.59e+02[4]($-$) & 2.48e+02[3] \\ 
     &  & (1.20e+03) & (1.28e+03) & (5.04e+02) & (1.49e+00) & (1.06e+01) & (1.24e+03) & (1.42e+02) \\  \hline
    \multirow{2}{*}{YLLF07} & \multirow{2}{*}{3.93e-28} & 1.62e+00[6]($-$) & 2.09e+00[7]($-$) & 6.67e-01[5]($-$) & 2.39e-01[2]($\approx$) & 2.90e-01[3]($-$) & 3.23e-01[4]($-$) & \hl{2.18e-01[1]} \\ 
     &  & (6.91e-01) & (9.17e-01) & (2.75e-01) & (9.54e-02) & (9.39e-02) & (1.56e-01) & (1.16e-01) \\  \hline
    \multirow{2}{*}{YLLF08} & \multirow{2}{*}{9.82e-28} & 5.71e+03[7]($-$) & 5.52e+03[5]($-$) & 4.66e+03[2]($-$) & 4.96e+03[4]($-$) & 4.70e+03[3]($-$) & 5.67e+03[6]($-$) & \hl{2.01e+03[1]} \\ 
     &  & (3.16e+02) & (2.88e+02) & (3.32e+02) & (2.73e+02) & (7.05e+02) & (2.83e+02) & (4.13e+02) \\  \hline
    \multirow{2}{*}{YLLF09} & \multirow{2}{*}{6.18e-23} & 1.84e+02[7]($-$) & 1.02e+02[2]($\approx$) & 1.57e+02[5]($-$) & 1.65e+02[6]($-$) & 1.34e+02[4]($-$) & 1.16e+02[3]($\approx$) & \hl{9.53e+01[1]} \\ 
     &  & (2.29e+01) & (2.09e+01) & (1.31e+01) & (1.68e+01) & (2.06e+01) & (5.14e+01) & (1.72e+01) \\  \hline
    \multirow{2}{*}{YLLF10} & \multirow{2}{*}{3.13e-33} & 1.61e+01[6]($-$) & 1.45e+01[5]($-$) & 1.33e+01[4]($-$) & 2.10e+00[3]($-$) & \hl{1.23e+00[1]($+$)} & 1.99e+01[7]($-$) & 1.52e+00[2] \\ 
     &  & (1.90e+00) & (1.34e+00) & (9.54e-01) & (3.09e-01) & (5.35e-01) & (2.80e-01) & (5.08e-01) \\  \hline
    \multirow{2}{*}{YLLF11} & \multirow{2}{*}{3.14e-32} & 5.48e+01[7]($-$) & 4.90e+01[6]($-$) & 3.14e+01[5]($-$) & \hl{1.01e+00[1]($+$)} & 1.17e+00[2]($+$) & 2.98e+01[4]($-$) & 3.28e+00[3] \\ 
     &  & (1.16e+01) & (1.24e+01) & (7.17e+00) & (1.63e-02) & (9.31e-02) & (2.00e+01) & (1.53e+00) \\  \hline
    \multirow{2}{*}{YLLF12} & \multirow{2}{*}{3.75e-27} & 6.01e+05[6]($-$) & 2.46e+06[7]($-$) & 6.09e+04[3]($-$) & 1.73e+05[4]($-$) & 4.29e+05[5]($-$) & 1.23e+02[2]($-$) & \hl{3.38e+01[1]} \\ 
     &  & (7.89e+05) & (3.04e+06) & (8.76e+04) & (3.15e+05) & (5.69e+05) & (5.49e+02) & (3.65e+01) \\  \hline
    \multirow{2}{*}{YLLF13} & \multirow{2}{*}{3.80e-34} & 5.47e+06[3]($+$) & 1.08e+07[4]($+$) & 1.10e+06[2]($+$) & 2.83e+10[6]($-$) & 4.76e+10[7]($-$) & \hl{6.82e+04[1]($+$)} & 7.79e+09[5] \\ 
     &  & (2.96e+06) & (8.07e+06) & (5.84e+05) & (1.12e+10) & (3.08e+10) & (1.94e+05) & (1.31e+10) \\  \hline
    mean rank &  & 6.00 & 5.41 & 4.35 & 2.41 & 3.47 & 4.06 & 2.29 \\ 
    $+$ / $-$ / $\approx$ &  & 1/16/0 & 2/14/1 & 1/16/0 & 7/7/3 & 7/10/0 & 2/12/3 & 0/0/0 \\ 
    \bottomrule
  \end{tabular}
  \label{tab:compar20d}
  \end{table*}

\subsection{Comparison study}
\label{subsec:comparison}

Table~\ref{tab:compar20d} presents the statistical results of seven optimization algorithms evaluated on two test suites. The results are presented in terms of p-values obtained from the Friedman test, mean ranks, and the corresponding Wilcoxon rank sum test. The highest rank in each row is denoted by grey shading, along with the corresponding ranks enclosed in brackets of each result. The p-value obtained from the Friedman test is considerably lower than 0.05, signifying a substantial difference between the outcomes. The analysis demonstrates that DRSO achieves the best mean rank of 2.29 out of seven algorithms across 17 test instances. The Skopt algorithm secured second position, and GPEME ranked third. Although EDA/LS is not primarily designed for expensive optimization, it still displays a strong competitive performance due to the use of VWH and local search methods. The FCPS-CoDE algorithm selects $N$ solutions for evaluation at each iteration, but its advantages are limited by the 500 evaluations allowed, nevertheless, it still outperforms the CMA-ES algorithm. According to the Wilcoxon rank sum test results, compared to DRSO, the most competitive algorithm, Skopt, achieved 7 better results, 7 worse results, and 3 results roughly equivalent. In low-dimensional problems, DRSO, driven by the relational model, has achieved statistical results similar to the most advanced BO algorithms, and DRSO even has an advantage in mean rank. Therefore, based on the aforementioned analysis, the DRSO algorithm demonstrates the best overall performance in the 20-dimensional search space.

\begin{table*}[ht!]
\renewcommand{\arraystretch}{1.1}
\renewcommand{\tabcolsep}{4pt}
\centering
\caption{Statistics of mean and standard deviation results obtained by five comparison algorithms on LZG and YLL test suites with $n=50$.} \scriptsize
\begin{tabular}{ccccccc}
\toprule
problem & p-value & CMA-ES & EDA/LS & SAMSO & FCPS-CoDE & DRSO \\
\midrule
\multirow{2}{*}{Ellipsoid} & \multirow{2}{*}{4.04e-22} & 1.91e+03[5]($-$) & 1.52e+03[3]($-$) & 1.31e+03[2]($\approx$) & 1.61e+03[4]($-$) & \hl{6.66e+02[1]} \\ 
 &  & (2.77e+02) & (2.23e+02) & (1.13e+03) & (3.39e+02) & (1.19e+02) \\  \hline
\multirow{2}{*}{Rosenbrock} & \multirow{2}{*}{1.88e-17} & 2.09e+03[5]($-$) & 1.78e+03[3]($-$) & 1.58e+03[2]($\approx$) & 1.90e+03[4]($-$) & \hl{8.81e+02[1]} \\ 
 &  & (4.12e+02) & (2.84e+02) & (1.70e+03) & (5.11e+02) & (1.66e+02) \\  \hline
\multirow{2}{*}{Ackley} & \multirow{2}{*}{5.24e-29} & 1.85e+01[4]($-$) & 1.76e+01[3]($-$) & 1.86e+01[5]($-$) & 1.75e+01[2]($-$) & \hl{1.45e+01[1]} \\ 
 &  & (9.68e-01) & (4.24e-01) & (1.18e+00) & (6.00e-01) & (5.64e-01) \\  \hline
\multirow{2}{*}{Griewank} & \multirow{2}{*}{3.37e-22} & 2.38e+02[2]($-$) & 2.41e+02[3]($-$) & 6.19e+02[5]($-$) & 2.69e+02[4]($-$) & \hl{1.12e+02[1]} \\ 
 &  & (3.80e+01) & (2.86e+01) & (3.66e+02) & (6.72e+01) & (2.26e+01) \\  \hline
\multirow{2}{*}{YLLF01} & \multirow{2}{*}{7.52e-24} & 2.61e+04[2]($-$) & 2.77e+04[3]($-$) & 6.69e+04[5]($-$) & 2.95e+04[4]($-$) & \hl{1.24e+04[1]} \\ 
 &  & (3.48e+03) & (3.92e+03) & (3.70e+04) & (5.86e+03) & (2.56e+03) \\  \hline
\multirow{2}{*}{YLLF02} & \multirow{2}{*}{3.17e-33} & 8.62e+13[4]($-$) & 2.79e+04[3]($-$) & 1.05e+18[5]($-$) & 8.92e+01[2]($-$) & \hl{6.54e+01[1]} \\ 
 &  & (3.40e+14) & (9.04e+04) & (3.77e+18) & (8.24e+00) & (7.20e+00) \\  \hline
\multirow{2}{*}{YLLF03} & \multirow{2}{*}{4.41e-21} & 1.36e+05[3]($\approx$) & 1.58e+05[4]($-$) & 2.97e+05[5]($-$) & \hl{8.24e+04[1]($+$)} & 1.34e+05[2] \\ 
 &  & (2.57e+04) & (2.51e+04) & (9.31e+04) & (1.68e+04) & (2.25e+04) \\  \hline
\multirow{2}{*}{YLLF04} & \multirow{2}{*}{5.21e-22} & 9.16e+01[5]($-$) & \hl{5.80e+01[1]($\approx$)} & 8.74e+01[4]($-$) & 5.88e+01[2]($\approx$) & 5.96e+01[3] \\ 
 &  & (1.26e+01) & (2.78e+00) & (8.06e+00) & (5.58e+00) & (4.95e+00) \\  \hline
\multirow{2}{*}{YLLF05} & \multirow{2}{*}{1.86e-25} & 3.78e+07[3]($-$) & 2.93e+07[2]($-$) & 1.55e+08[5]($-$) & 4.49e+07[4]($-$) & \hl{1.16e+07[1]} \\ 
 &  & (1.25e+07) & (5.30e+06) & (8.04e+07) & (1.82e+07) & (3.71e+06) \\  \hline
\multirow{2}{*}{YLLF06} & \multirow{2}{*}{1.67e-28} & 2.82e+04[3]($-$) & 2.74e+04[2]($-$) & 7.82e+04[5]($-$) & 2.92e+04[4]($-$) & \hl{1.19e+04[1]} \\ 
 &  & (5.82e+03) & (4.41e+03) & (3.06e+04) & (7.35e+03) & (2.11e+03) \\  \hline
\multirow{2}{*}{YLLF07} & \multirow{2}{*}{1.42e-25} & 3.14e+01[4]($-$) & 2.17e+01[2]($-$) & 1.31e+02[5]($-$) & 2.96e+01[3]($-$) & \hl{8.80e+00[1]} \\ 
 &  & (9.15e+00) & (4.75e+00) & (7.91e+01) & (1.28e+01) & (3.10e+00) \\  \hline
\multirow{2}{*}{YLLF08} & \multirow{2}{*}{2.68e-28} & 1.65e+04[4]($-$) & 1.51e+04[2]($-$) & 1.66e+04[5]($-$) & 1.63e+04[3]($-$) & \hl{1.38e+04[1]} \\ 
 &  & (5.98e+02) & (5.84e+02) & (6.62e+02) & (6.11e+02) & (6.19e+02) \\  \hline
\multirow{2}{*}{YLLF09} & \multirow{2}{*}{6.58e-23} & 6.21e+02[5]($-$) & 5.19e+02[4]($-$) & 4.97e+02[3]($\approx$) & \hl{3.80e+02[1]($+$)} & 4.82e+02[2] \\ 
 &  & (4.52e+01) & (1.78e+01) & (1.11e+02) & (3.15e+01) & (2.39e+01) \\  \hline
\multirow{2}{*}{YLLF10} & \multirow{2}{*}{2.76e-30} & 1.89e+01[4]($-$) & 1.75e+01[3]($-$) & 1.94e+01[5]($-$) & 1.72e+01[2]($-$) & \hl{4.78e+00[1]} \\ 
 &  & (1.45e+00) & (3.88e-01) & (2.95e+00) & (8.67e-01) & (2.54e-01) \\  \hline
\multirow{2}{*}{YLLF11} & \multirow{2}{*}{1.13e-21} & 2.47e+02[3]($-$) & 2.42e+02[2]($-$) & 6.31e+02[5]($-$) & 2.52e+02[4]($-$) & \hl{1.10e+02[1]} \\ 
 &  & (3.82e+01) & (2.55e+01) & (3.41e+02) & (6.26e+01) & (2.07e+01) \\  \hline
\multirow{2}{*}{YLLF12} & \multirow{2}{*}{5.66e-22} & 4.07e+07[3]($-$) & 1.64e+07[2]($-$) & 5.10e+08[5]($-$) & 5.44e+07[4]($-$) & \hl{6.61e+06[1]} \\ 
 &  & (2.32e+07) & (5.98e+06) & (3.16e+08) & (3.49e+07) & (4.32e+06) \\  \hline
\multirow{2}{*}{YLLF13} & \multirow{2}{*}{8.00e-29} & 1.03e+08[2]($+$) & \hl{7.53e+07[1]($+$)} & 8.32e+08[4]($+$) & 1.58e+08[3]($+$) & 1.59e+12[5] \\ 
 &  & (3.96e+07) & (2.19e+07) & (6.39e+08) & (1.01e+08) & (5.80e+11) \\  \hline
mean rank &  & 3.59 & 2.53 & 4.41 & 3.00 & 1.47 \\ 
$+$ / $-$ / $\approx$ &  & 1/15/1 & 1/15/1 & 1/13/3 & 3/13/1 &  \\ 
\bottomrule
\end{tabular}
\label{tab:compar50d}
\end{table*}
The statistical results in the 50-dimensional problem are shown in Table~\ref{tab:compar50d}, and DRSO still shows the best performance, achieving an average rank of 1.47 among the five algorithms. Based on the results of the Wilcoxon rank sum test, the four comparative algorithms have 15, 14, 13, and 13 results inferior to DRSO out of 17 problems, respectively.

\subsection{Ablation study}
In this section, we will conduct ablation experiments on several important components of the DRSO, including the offspring selection strategy, the relation model, and the generation of new solutions. The details of the algorithm variants are shown in Table \ref{tab:abl_setting}.

\begin{table*}[ht!]
\renewcommand{\arraystretch}{1.1}
\renewcommand{\tabcolsep}{4pt}
\centering
\caption{Design of algorithm variants.} \scriptsize
\begin{tabular}{c|c}
\toprule
algorithm & details \\
\midrule
DRSO & Utilizes default settings \\ 
DRSO-Sel-1 & Selects $\mathcal{P}{u}$ at random, $\mathcal{M}{2}$ model excluded \\ 
DRSO-Sel-2 & Selects $\mathcal{Q}{best}$ at random, $\mathcal{M}_{1}$ model excluded \\ 
DRSO-Gen-1 & Employs $\mathcal{P}{e}$ for new solutions in EDA, without $\mathcal{P}{u}$\\ 
DRSO-Gen-2 & Excludes local search in the improvement of solution quality \\ 
DRSO-Mod & Excludes relation model, employs only XGBoost for classification and regression, selects $\mathcal{P}{u}$ and $\mathcal{Q}{best}$ \\ 
\bottomrule
\end{tabular}
\label{tab:abl_setting}
\end{table*}

DRSO-Sel-1 and DRSO-Sel-2 are used to verify the importance of reliably selecting $\mathcal{P}{u}$ and $\mathcal{Q}_{best}$. DRSO-Gen-1 and DRSO-Gen-2 serve to verify the significance of $\mathcal{P}{u}$ and the local search algorithm. DRSO-Mod is utilized to validate the effectiveness of the relation model. Experiments were independently conducted 30 times on LZG test suit with 20 and 50 dimensions. The experimental design and result statistics are consistent with the \ref{subsec:comparison} section. The experimental results are presented in Table \ref{tab:blation_result}.

\begin{table*}[htbp]
\renewcommand{\arraystretch}{1.1}
\renewcommand{\tabcolsep}{4pt}
\centering
\caption{Ablation study results comparing DRSO and its five variants on LZG test suites with n=20,50.} \scriptsize
\begin{tabular}{ccccccccc}
\toprule
n & problem & p-value & DRSO-GEN-1 & DRSO-GEN-2 & DRSO-SEL-1 & DRSO-SEL-2 & DRSO-MODEL & DRSO \\
\midrule
\multirow{8}{*}{20}&\multirow{2}{*}{Ellipsoid} & \multirow{2}{*}{1.13e-15} & 3.36e+01[5]($-$) & 1.18e+01[3]($-$) & 7.24e+00[2]($\approx$) & 4.09e+01[6]($-$) & 1.39e+01[4]($-$) & \hl{6.17e+00[1]} \\ 
& &  & (3.73e+01) & (5.15e+00) & (4.82e+00) & (1.33e+01) & (7.24e+00) & (4.57e+00) \\  \cline{2-9}
& \multirow{2}{*}{Rosenbrock} & \multirow{2}{*}{5.04e-05} & 1.27e+02[5]($-$) & 1.05e+02[3]($\approx$) & \hl{9.54e+01[1]($\approx$)} & 1.39e+02[6]($-$) & 1.24e+02[4]($\approx$) & 1.02e+02[2] \\ 
& &  & (4.79e+01) & (3.19e+01) & (3.26e+01) & (3.35e+01) & (4.50e+01) & (2.94e+01) \\  \cline{2-9}
& \multirow{2}{*}{Ackley} & \multirow{2}{*}{6.18e-21} & 8.71e+00[4]($-$) & 7.86e+00[3]($-$) & \hl{5.98e+00[1]($\approx$)} & 1.12e+01[6]($-$) & 9.94e+00[5]($-$) & 6.08e+00[2] \\ 
& & & (2.52e+00) & (9.18e-01) & (9.91e-01) & (9.06e-01) & (1.30e+00) & (1.05e+00) \\  \cline{2-9}
& \multirow{2}{*}{Griewank} & \multirow{2}{*}{1.06e-19} & 1.31e+01[5]($-$) & 6.45e+00[3]($-$) & 3.65e+00[2]($\approx$) & 1.65e+01[6]($-$) & 8.57e+00[4]($-$) & \hl{3.08e+00[1]} \\ 
&  & & (1.67e+01) & (2.09e+00) & (1.49e+00) & (4.59e+00) & (3.01e+00) & (1.54e+00) \\  \hline
\multirow{8}{*}{50}&\multirow{2}{*}{Ellipsoid} & \multirow{2}{*}{9.52e-22} & 7.61e+02[3]($-$) & 1.28e+03[6]($-$) & 6.97e+02[2]($\approx$) & 1.05e+03[4]($-$) & 1.13e+03[5]($-$) & \hl{6.66e+02[1]} \\ 
& &  & (2.08e+02) & (2.01e+02) & (9.68e+01) & (1.65e+02) & (2.08e+02) & (1.19e+02) \\  \cline{2-9}
&\multirow{2}{*}{Rosenbrock} & \multirow{2}{*}{1.51e-21} & 1.02e+03[3]($-$) & 2.01e+03[6]($-$) & 9.69e+02[2]($\approx$) & 1.13e+03[4]($-$) & 1.54e+03[5]($-$) & \hl{8.81e+02[1]} \\ 
&&  & (2.10e+02) & (3.02e+02) & (1.90e+02) & (2.18e+02) & (2.88e+02) & (1.66e+02) \\  \cline{2-9}
&\multirow{2}{*}{Ackley} & \multirow{2}{*}{4.03e-23} & 1.50e+01[3]($-$) & 1.73e+01[6]($-$) & 1.49e+01[2]($-$) & 1.59e+01[4]($-$) & 1.71e+01[5]($-$) & \hl{1.45e+01[1]} \\ 
&&  & (6.13e-01) & (4.31e-01) & (5.92e-01) & (6.57e-01) & (5.49e-01) & (5.64e-01) \\  \cline{2-9}
&\multirow{2}{*}{Griewank} & \multirow{2}{*}{4.37e-22} & 1.26e+02[3]($\approx$) & 2.06e+02[6]($-$) & 1.16e+02[2]($\approx$) & 1.70e+02[4]($-$) & 1.98e+02[5]($-$) & \hl{1.12e+02[1]} \\ 
& &  & (3.32e+01) & (2.90e+01) & (1.58e+01) & (2.61e+01) & (2.95e+01) & (2.26e+01) \\  \hline
& mean rank &  & 3.875 & 4.50 & 1.75 & 5.00 & 4.625 & 1.25 \\ 
& $+$ / $-$ / $\approx$ &  & 0/7/1 & 0/7/1 & 0/1/7 & 0/8/0 & 0/7/1 &  \\ 
\bottomrule
\end{tabular}
\label{tab:blation_result}
\end{table*}

Broadly speaking, all the algorithmic variants that omit a particular module are inferior to the original algorithm in terms of results. Specifically, the results of DRSO-GEN-1 and DRSO-GEN-2 are significantly inferior to the original version on 7 problems, indicating the importance of $\mathcal{P}_{u}$ and local search in generating solutions. The results of DRSO-SEL-2 are also poor, being inferior to the original algorithm on all problems, highlighting the importance of the reliable selection of $\mathcal{Q}_{best}$. The performance of DRSO-MODEL is significantly worse than DRSO on 7 problems, demonstrating the significance of the relation model. The performance deterioration of DRSO-SEL-1 is not obvious, as it is inferior to the original algorithm on only one problem, but its mean rank is still worse than the DRSO, indicating that the contribution of $\mathcal{P}_{u}$ to the algorithm is not as significant as $\mathcal{Q}_{best}$. In summary, each component of DRSO is effective, and their synergistic effect is also effective.

\subsection{Analysis of relation model}

To analyze the fitting capacity of the relation model, four representative functions from the LZG test suite were chosen. The fitting ability of the relation model was visualized in a two-dimensional search space. Additionally, a comparison was made between the relation model's capability and that of regression and classification models in selecting $\mathcal{Q}_{best}$ and $\mathcal{P}_u$ in 20-dimensional problems. The results demonstrate the advantages of the relation model in model-assisted selection.

\subsubsection{Visualization analysis}
\begin{figure}[ht!]
  \centering
  \subfloat[Ellipsoid]{
  \begin{minipage}{0.12\textwidth}
  \centering
  \includegraphics[width=1\columnwidth]{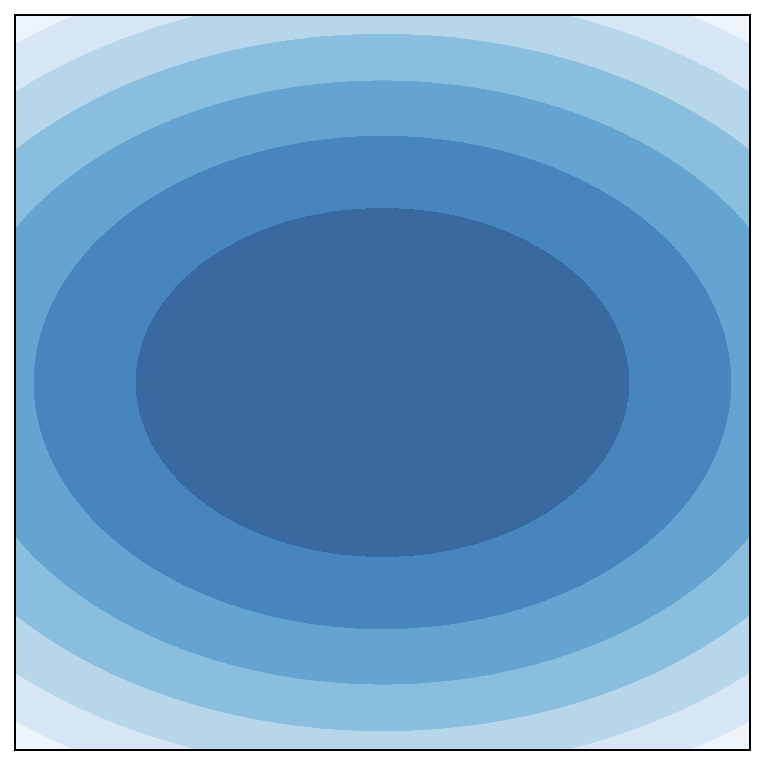}
  \includegraphics[width=1\columnwidth]{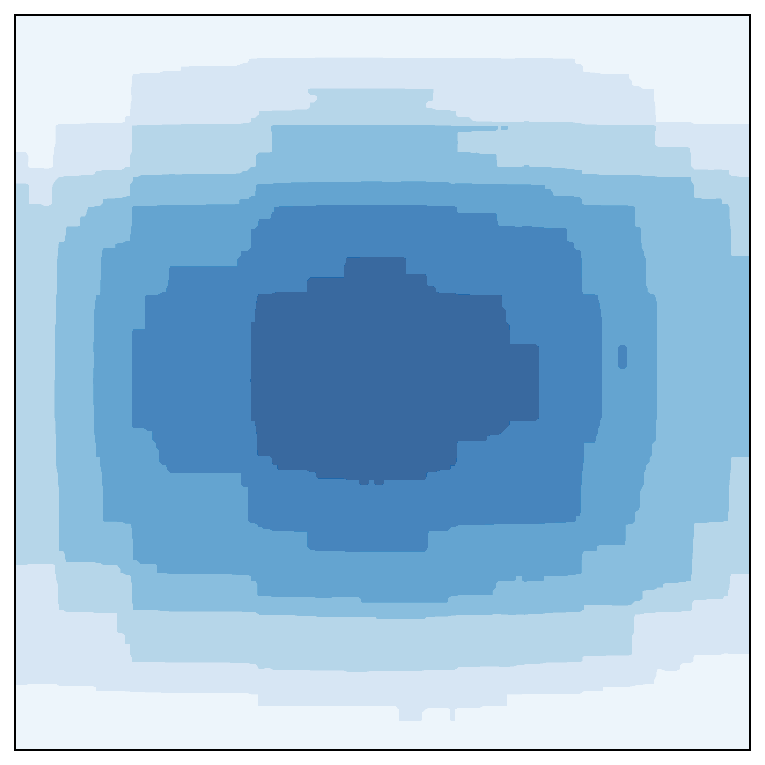}
  \end{minipage}}
  \subfloat[Rosenbrock]{
  \begin{minipage}{0.12\textwidth}
  \centering
  \includegraphics[width=1\columnwidth]{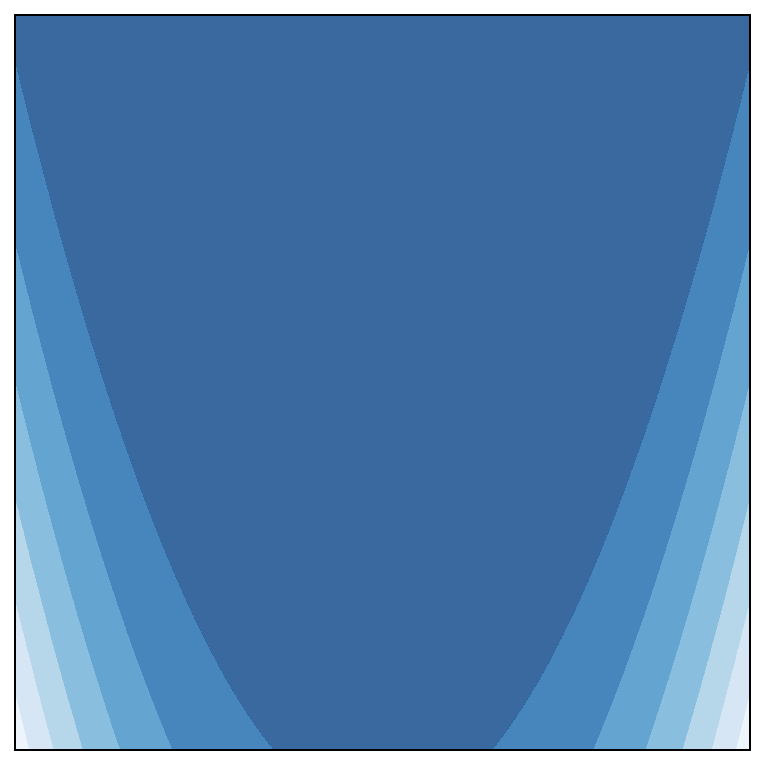}
  \includegraphics[width=1\columnwidth]{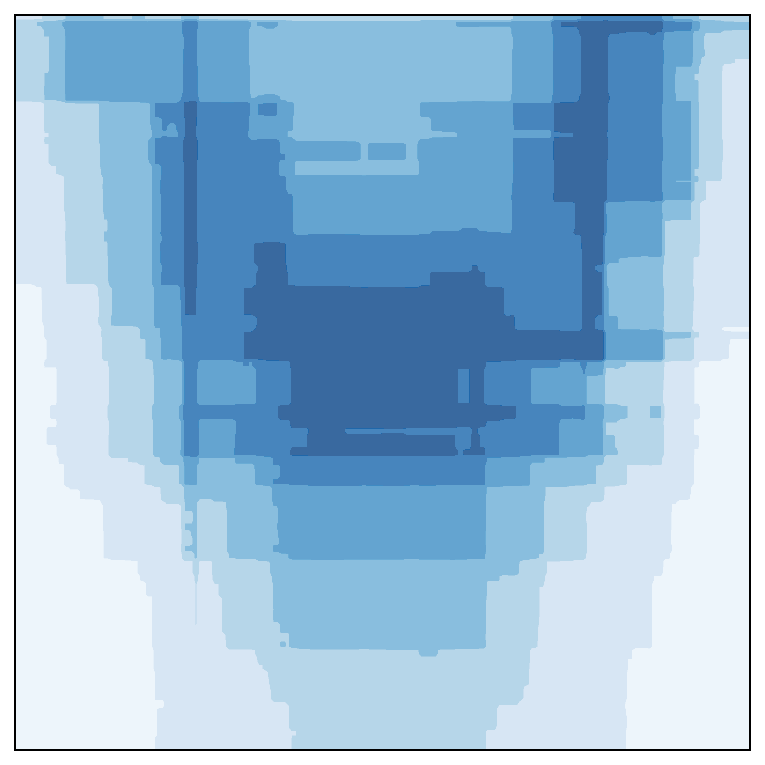}
  \end{minipage}}
  \subfloat[Ackley]{
  \begin{minipage}{0.12\textwidth}
  \centering
  \includegraphics[width=1\columnwidth]{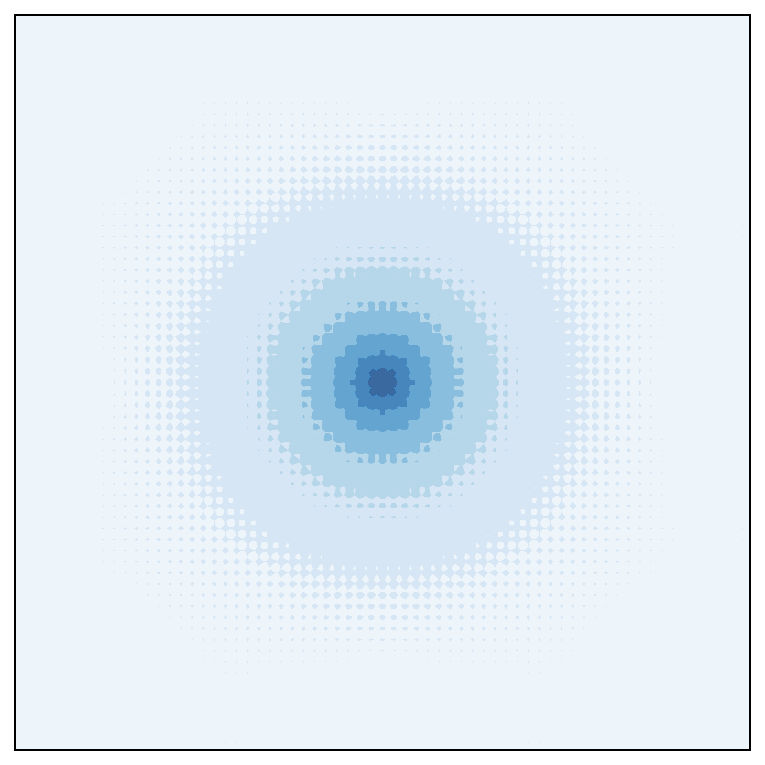}
  \includegraphics[width=1\columnwidth]{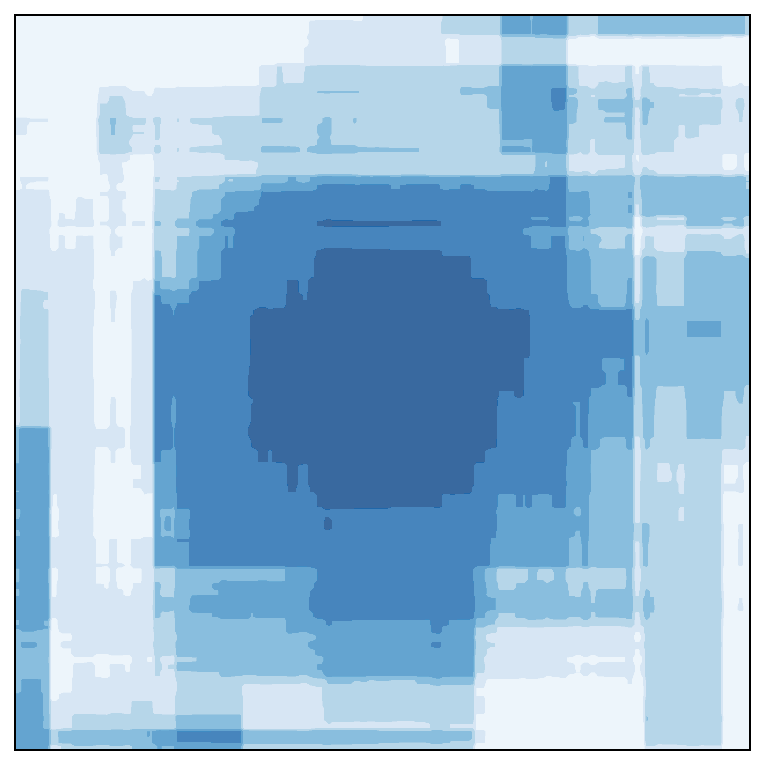} 
  \end{minipage}}
  \subfloat[Griewank]{
  \begin{minipage}{0.12\textwidth}
  \centering
  \includegraphics[width=1\columnwidth]{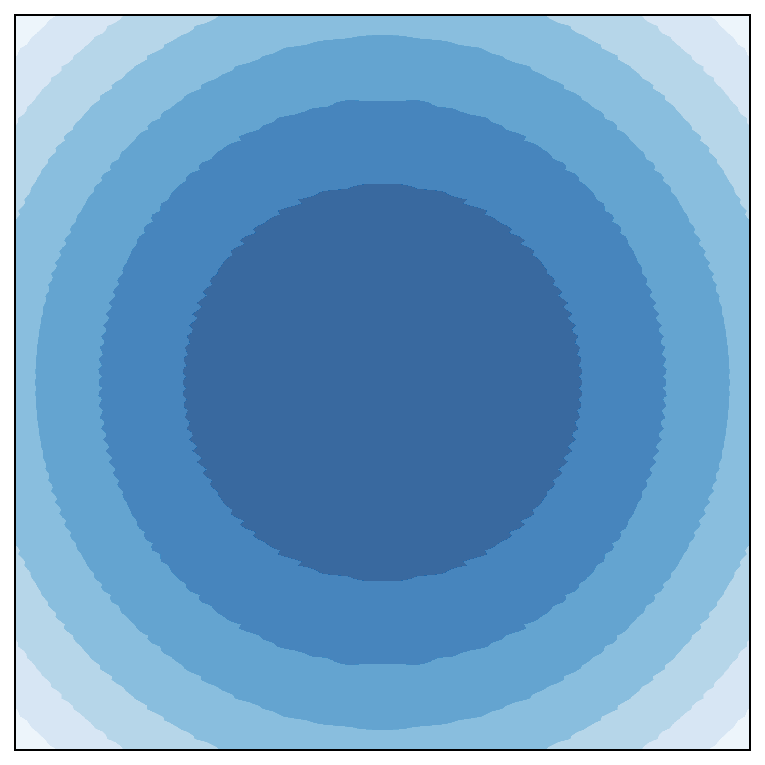}
  \includegraphics[width=1\columnwidth]{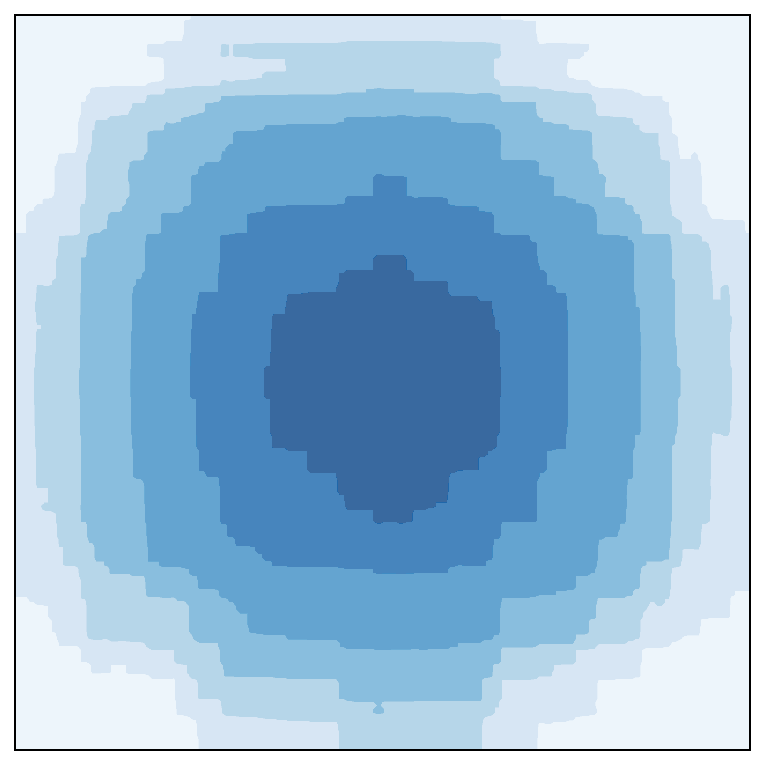}
  \end{minipage}}
  \caption{Contour plot of predicted results for the C1 criterion relation model in 2-dimensional space. The first row shows results based on real function values, while the second row shows predicted results.}
  \label{fig:C1_plot}
  \end{figure}

In the first row of Figure~\ref{fig:C1_plot}, the contour distributions of the four test functions are depicted in their corresponding search spaces. Following this, LHS sampling was utilized to generate 100 points from the space as the original training data. Subsequently, a relation model~($\mathcal{M}_{1}$) is then constructed using the C1 criterion on training data. Based on the predicted values from the $\mathcal{M}_{1}$ model, the contour distribution is displayed in the second row of Figure~\ref{fig:C1_plot}. It is apparent that the relation model, under the C1 criterion, resembles a regression-like process and is capable of acquiring knowledge on the original function's landscape. For instance, in the case of the Ellipsoid function, distinguished by its unimodal feature, and the Rosenbrock function, identified by a gully. the relation model does not intentionally fit the distribution of local extremums. However, it can still effectively represent the overall landscape of these two functions, which is vital for model-assisted search.

\begin{figure}[ht!]
\centering
\subfloat[Ellipsoid]{
\begin{minipage}{0.12\textwidth}
\centering
\includegraphics[width=1\columnwidth]{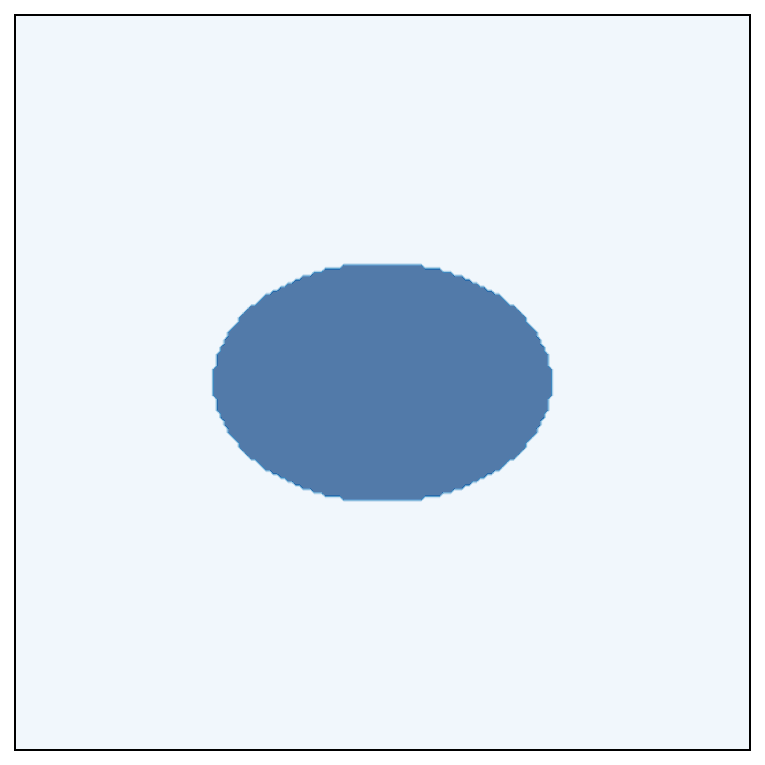}
\includegraphics[width=1\columnwidth]{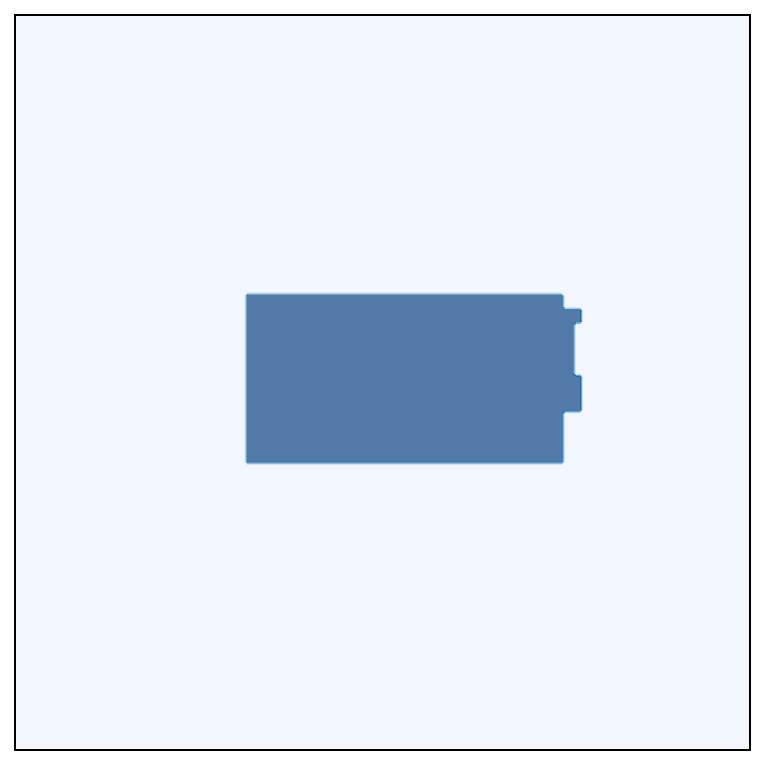}
\end{minipage}\label{fig:C2_plot_a}}
\subfloat[Rosenbrock]{
\begin{minipage}{0.12\textwidth}
\centering
\includegraphics[width=1\columnwidth]{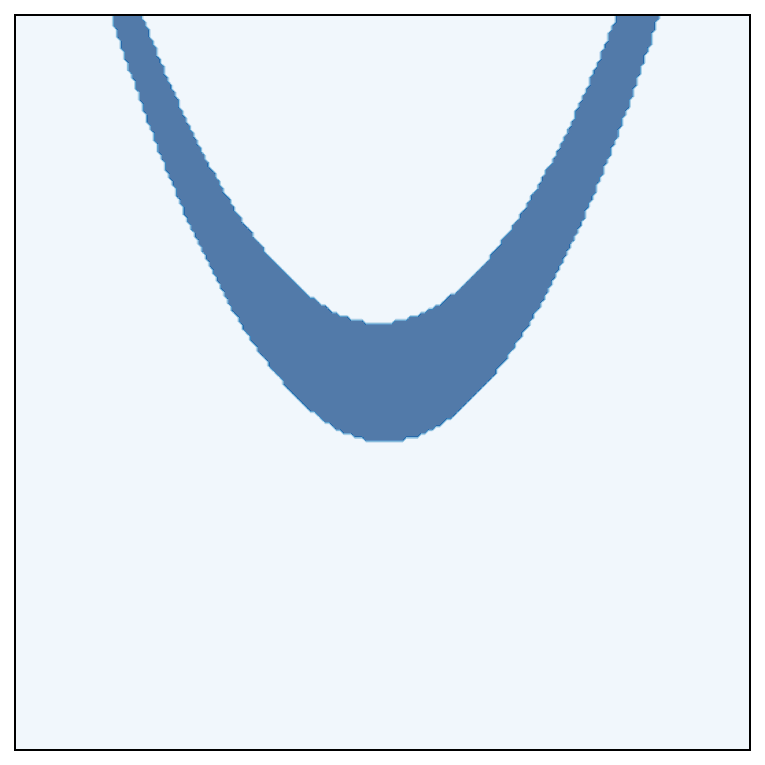}
\includegraphics[width=1\columnwidth]{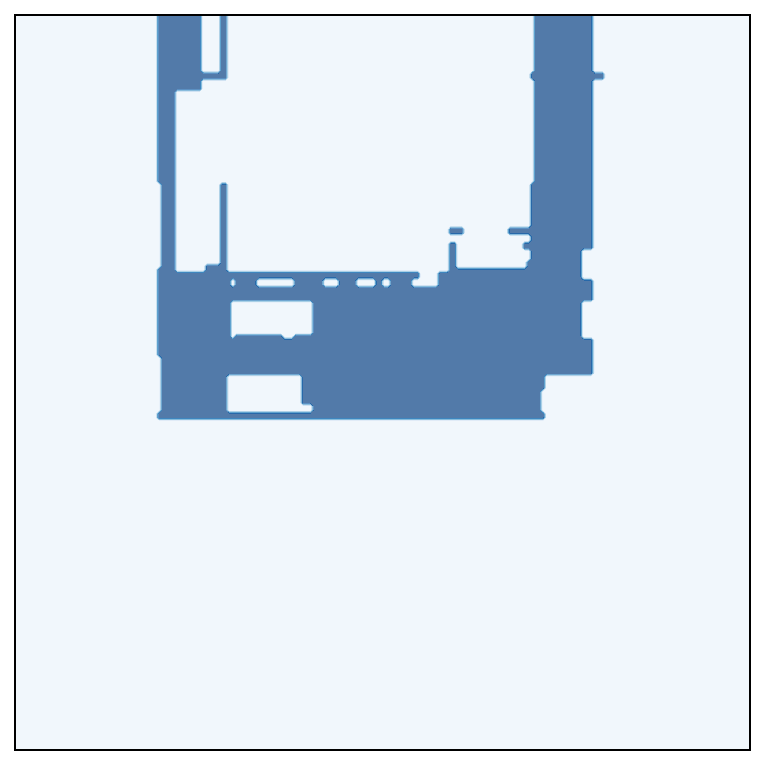}
\end{minipage}\label{fig:C2_plot_b}}
\subfloat[Ackley]{
\begin{minipage}{0.12\textwidth}
\centering
\includegraphics[width=1\columnwidth]{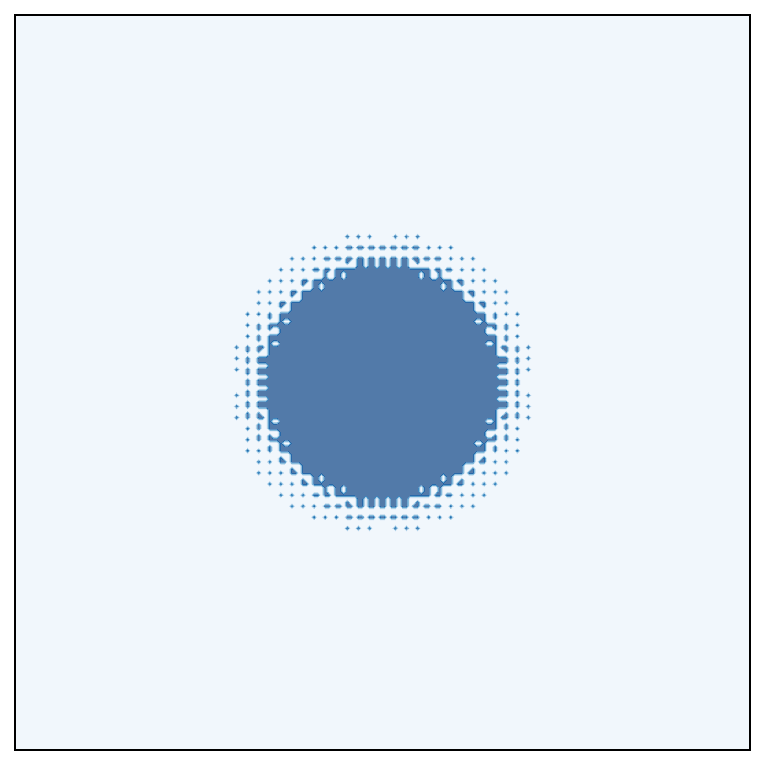}
\includegraphics[width=1\columnwidth]{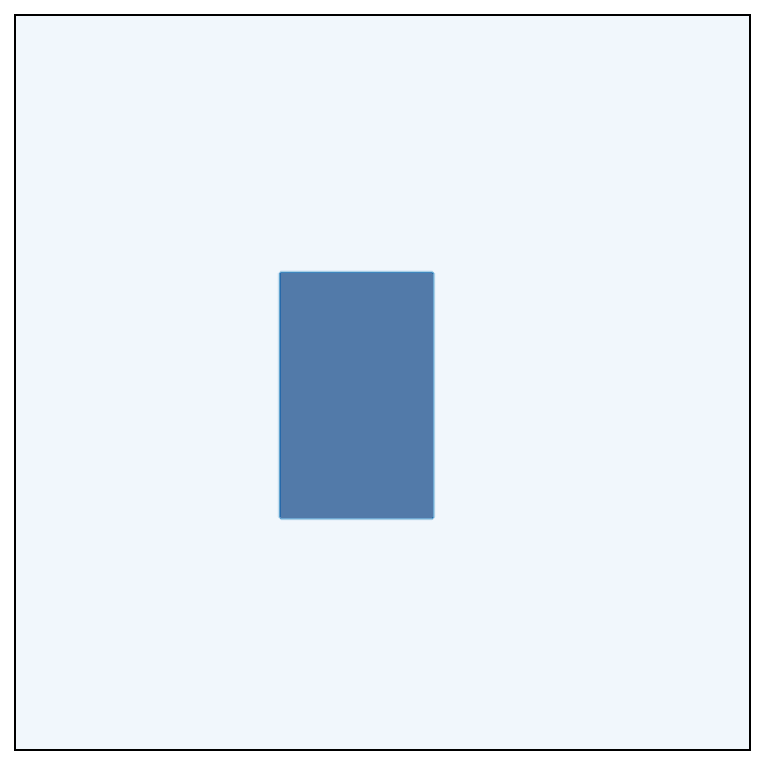} 
\end{minipage}\label{fig:C2_plot_c}}
\subfloat[Griewank]{
\begin{minipage}{0.12\textwidth}
\centering
\includegraphics[width=1\columnwidth]{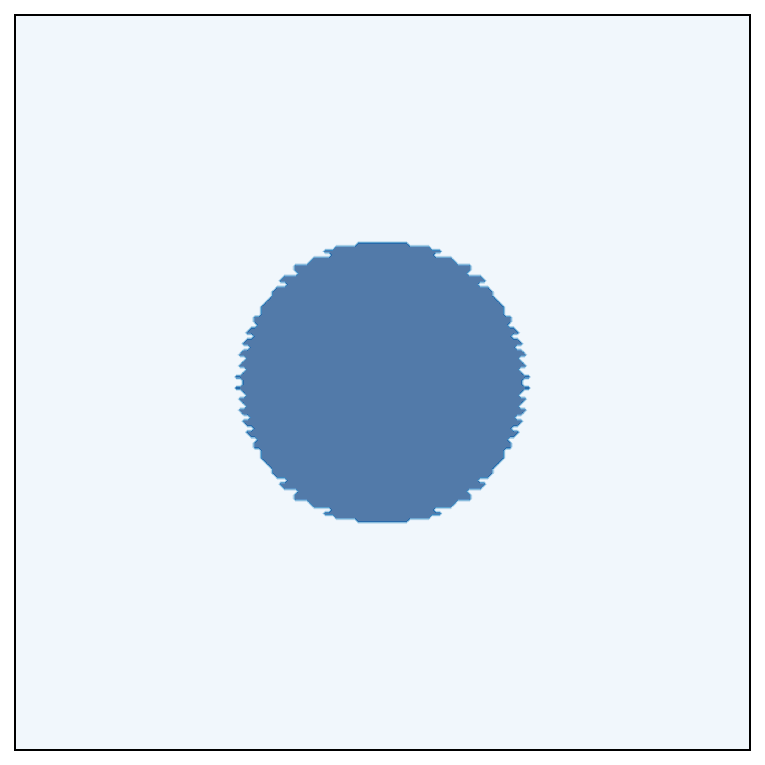}
\includegraphics[width=1\columnwidth]{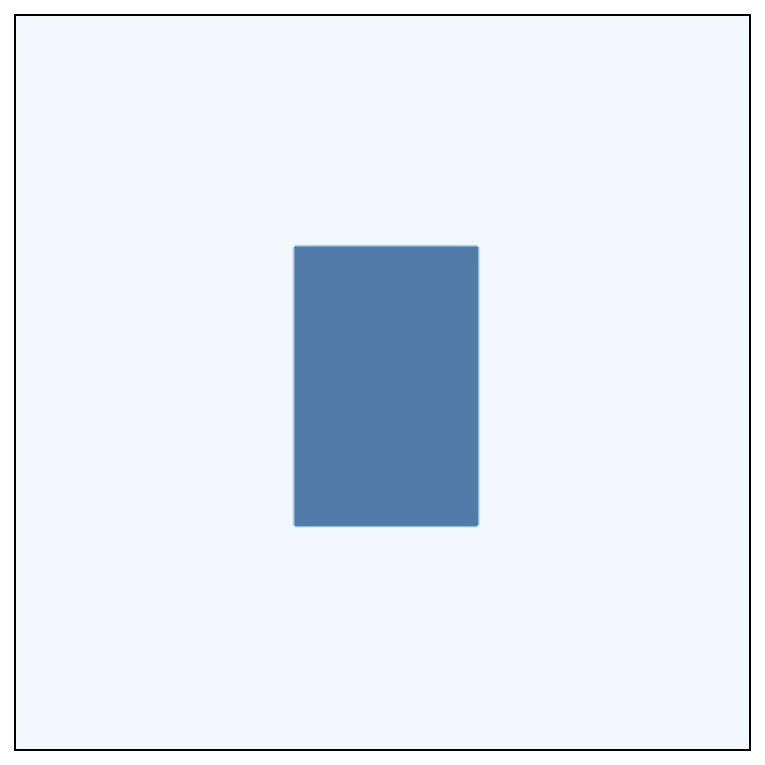}
\end{minipage}\label{fig:C2_plot_d}}
\subfloat[Ellipsoid]{
\begin{minipage}{0.12\textwidth}
\centering
\includegraphics[width=1\columnwidth]{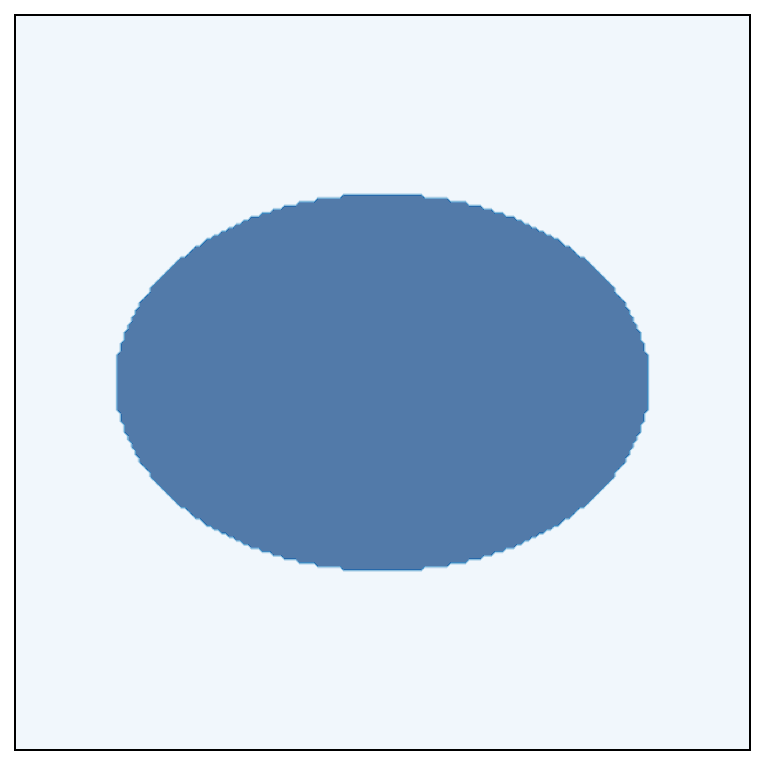}
\includegraphics[width=1\columnwidth]{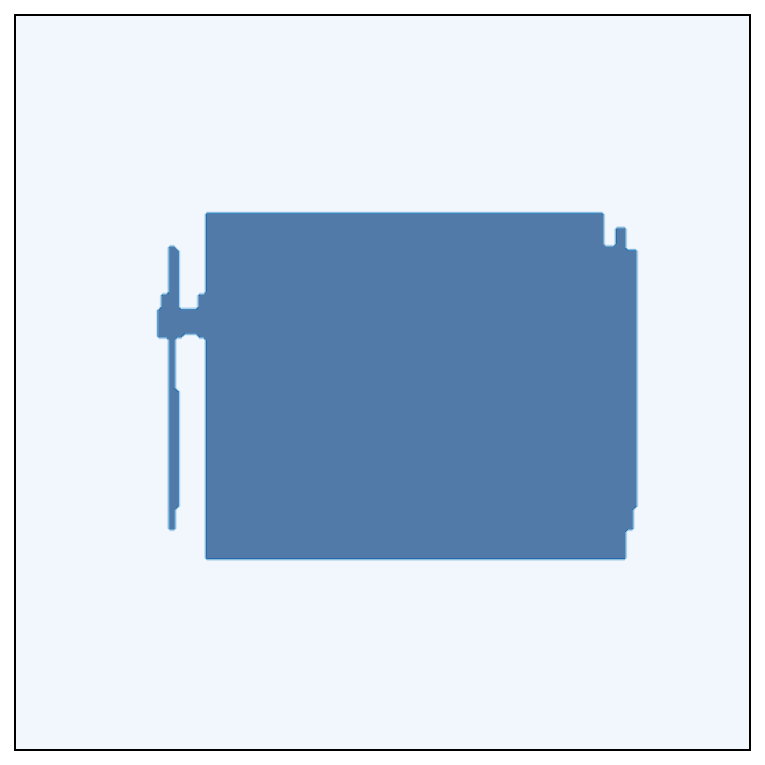}
\end{minipage}\label{fig:C2_plot_e}}
\subfloat[Rosenbrock]{
\begin{minipage}{0.12\textwidth}
\centering
\includegraphics[width=1\columnwidth]{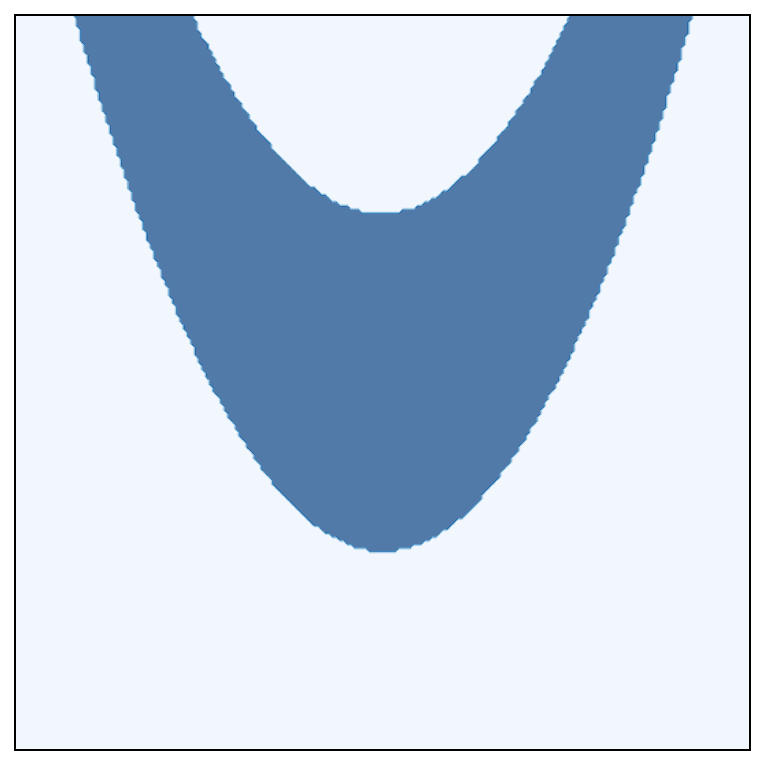}
\includegraphics[width=1\columnwidth]{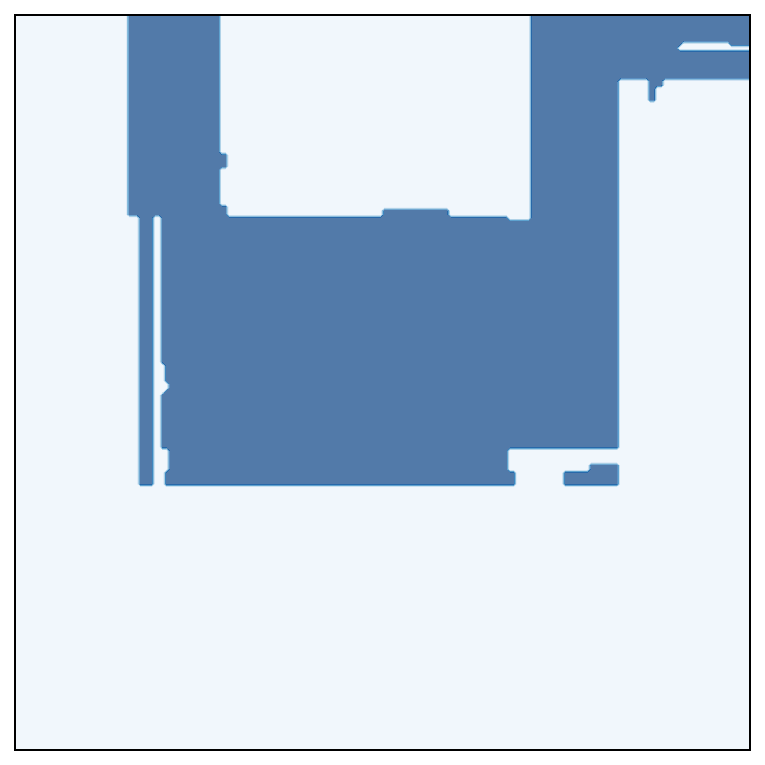}
\end{minipage}\label{fig:C2_plot_f}}
\subfloat[Ackley]{
\begin{minipage}{0.12\textwidth}
\centering
\includegraphics[width=1\columnwidth]{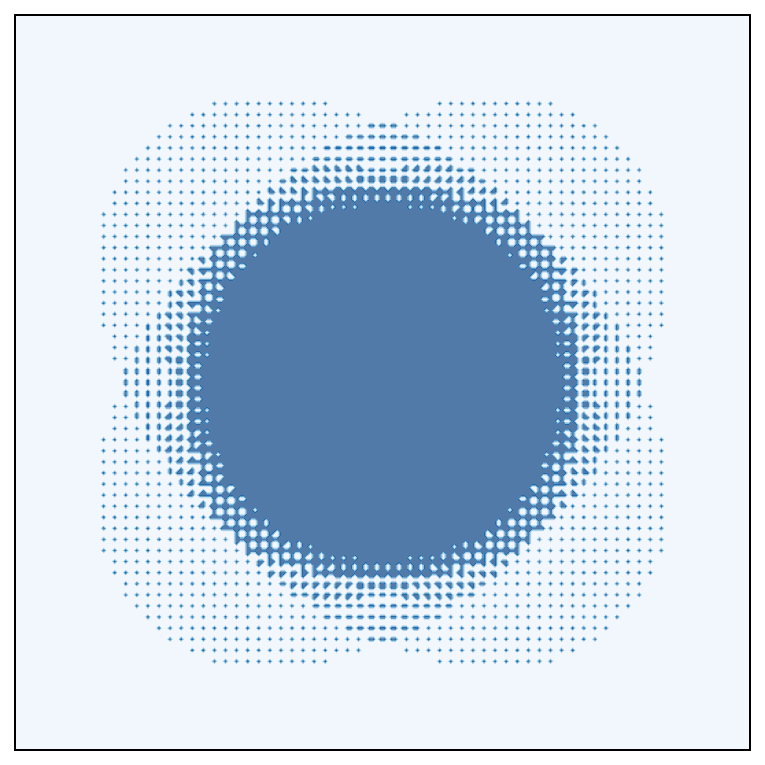}
\includegraphics[width=1\columnwidth]{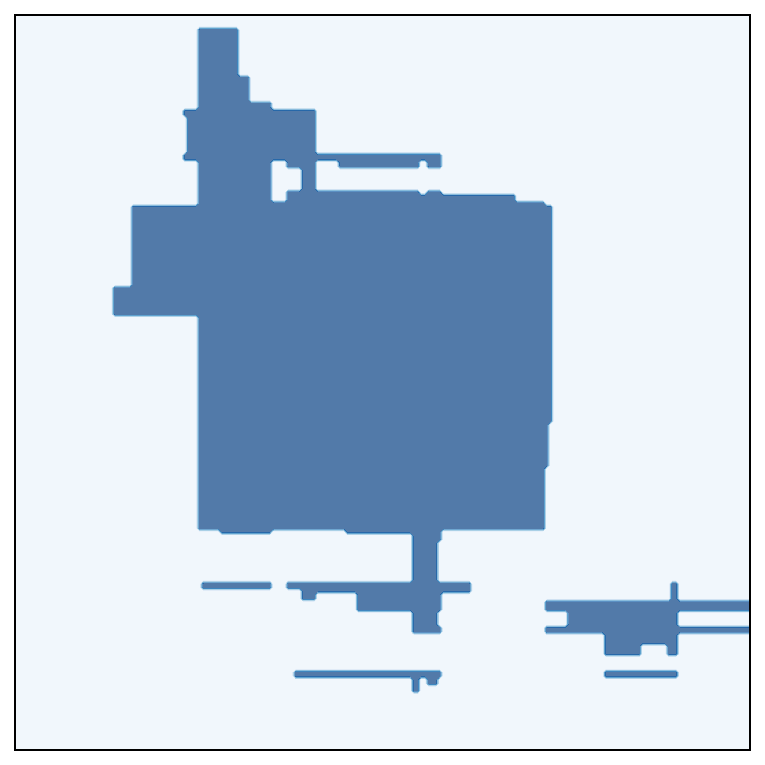}
\end{minipage}\label{fig:C2_plot_g}}
\subfloat[Griewank]{
\begin{minipage}{0.12\textwidth}
\centering
\includegraphics[width=1\columnwidth]{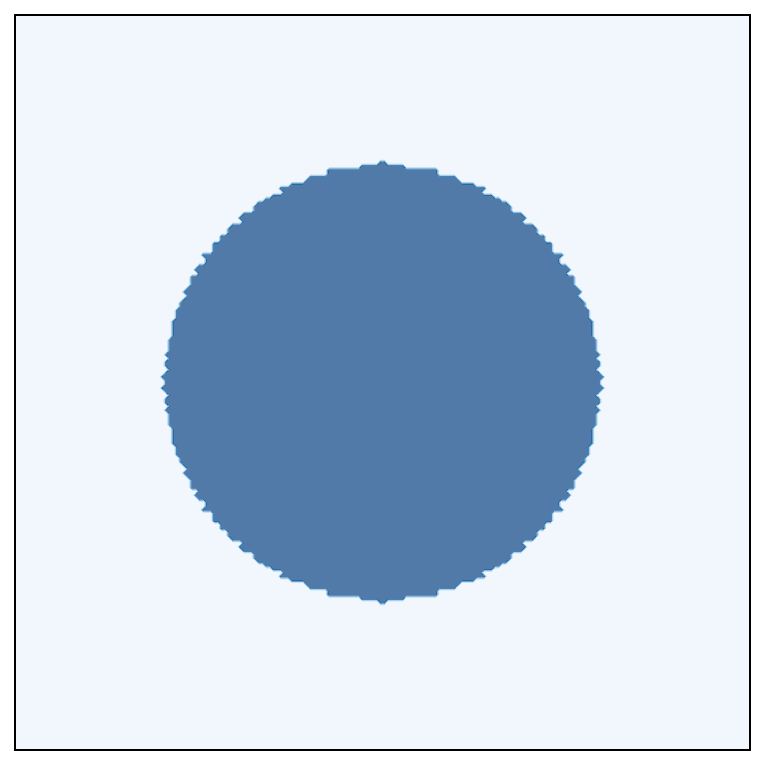}
\includegraphics[width=1\columnwidth]{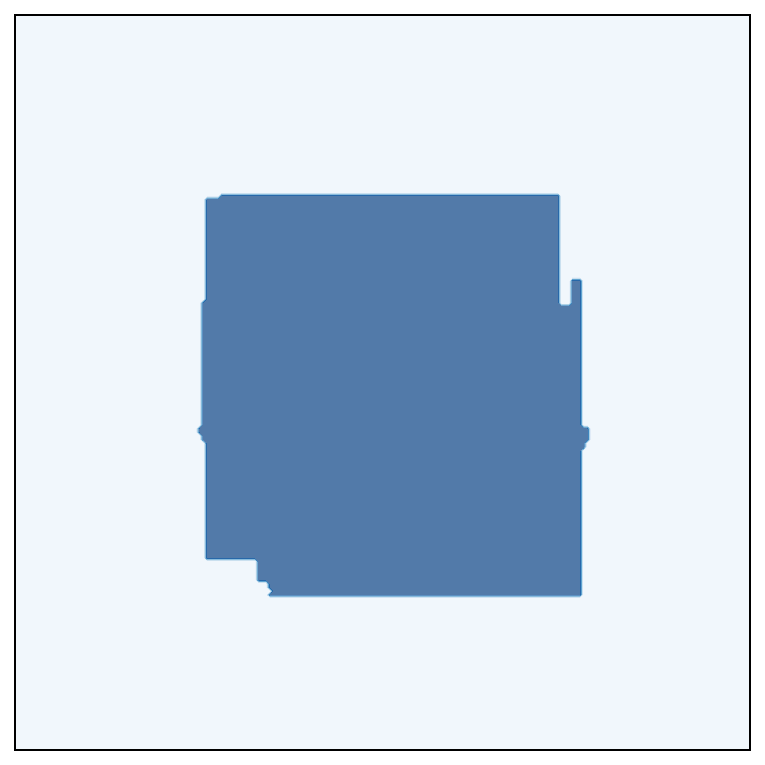}
\end{minipage}\label{fig:C2_plot_h}}
\caption{Contour plot of predicted results for the C2 criterion relation model in 2-dimensional space. The first row shows results based on true function values, while the second row shows predicted results. Fig (a)-(d) show results for $t=10\%$, while Fig (e)-(h) show results for $t=30\%$.}
\label{fig:C2_plot}
\end{figure}

In the first row of Figure~\ref{fig:C2_plot}, the distribution of data based on the true objective function and the threshold is depicted, where cases Figure~\ref{fig:C2_plot_a}-\ref{fig:C2_plot_d} correspond to a classification threshold of $t=10\%$, and cases Figure~\ref{fig:C2_plot_e}-\ref{fig:C2_plot_h} correspond to a threshold of $t=30\%$. A smaller threshold indicates a narrower focus area. LHS is utilized to extract 100 data points from the decision space. Subsequently, a relation model $\mathcal{M}_2$ is trained based on the C2 criterion and the specified threshold. The prediction outcomes are presented in the second row of Figure~\ref{fig:C2_plot}. Notably, the C2 criterion resembles a classification-like model, proficient in recognizing the classification boundaries of the original data and modifying the range of the model fitting boundary as per the threshold $t$. Additionally, the relation data's label balance strategy ensures that the model training remains unaffected by imbalanced class proportions, even when $t=10\%$.
  
\subsubsection{Accuracy analysis}

The relation model showcases properties akin to both classification and regression models. This raises a valid question - why not directly employ either a classification or regression model? In the subsequent analysis, we will explore the advantages of utilizing the relation model over classification and regression models in the context of model-assisted selection. 

To accentuate the importance of the data preparation and model usage stages in the relation model, we have excluded the differences in the learning abilities of the machine learning algorithms. We have opted for XGBoost with default parameters as the fundamental method for regression, classification, and the two relation models. These models are denoted as XGBR, XGBC, R-C1, and R-C2, respectively. To eliminate the randomness in the operation of EAs, the parent population $\mathcal{P}$ and offspring population $\mathcal{Q}$ information generated by GA in 50 consecutive generations on the 20-dimensional LZG test suits are stored. The population size $N$ is set to 50. The parent population is used as the training data for the model, while the offspring population is used as the testing data. To uniformly evaluate the capabilities of each model, two accuracy indicators, $acc_1$ and $acc_2$, are used to evaluate the performance on $\mathcal{Q}_{best}$ and $\mathcal{P}_{u}$, respectively. The calculation methods for $acc_1$ and $acc_2$ are as follows:
\begin{equation}
  \label{equ:acc1}
  acc_{1} = R(Q_{best}',\mathcal{Q})
  \end{equation}
where $\mathcal{Q}$ refers to both the offspring population and the test data. $\mathcal{Q}_{best}'$ denotes the best solution that is selected by the model within $\mathcal{Q}$. The function $R(\cdot)$ returns the ranking of $\mathcal{Q}_{best}'$ within the $\mathcal{Q}$ based on the real objective values. A smaller value of $acc_1$ indicates a higher effectiveness of the model in selecting the best solution.
\begin{equation}
\label{equ:acc2}
acc_{2} = \frac{ |\mathcal{P}_{u} \cap \mathcal{P}_{u}'|}{|\mathcal{P}_{u}|} 
\end{equation}
$\mathcal{P}_{u}$ represents the top $t$ of solutions selected based on the real objective values. $\mathcal{P}_{u}'$ denotes the selection made by the model, while $acc_{2}$ represents the proportion of cases where the model's selection matches the actual result. A higher value of $acc_{2}$ indicates a stronger ability of the model to make accurate selections.
\begin{figure}[htbp]
  \centering
  \includegraphics[width=\textwidth,trim=0 25 0 0,clip]{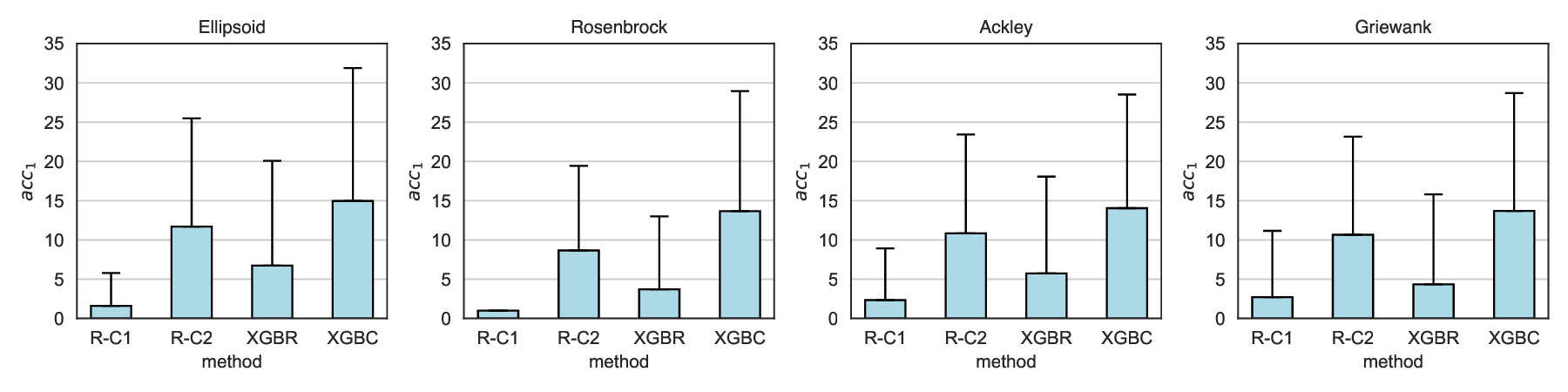} 
\caption{Bar chart of the $acc_1$ statistics of $\mathcal{Q}_{best}$ selections for different surrogate models.}
\label{fig:select_tops}
\end{figure}

Based on the results shown in Figure~\ref{fig:select_tops}, which is the bar of the $acc1$ metric for selecting $Q_{best}$ over 50 generations, it can be observed that the R-C1 performs the best across all problems, with the smallest average rank value and error bar. This suggests that the R-C1 is more suitable for scenarios where the top 1 solution needs to be selected from a set. The R-C2 performs worse than the regression model XGBR, but better than the classification model XGBC.
\begin{figure}[htbp]
  \centering
  \includegraphics[width=\textwidth,trim=0 25 0 0,clip]{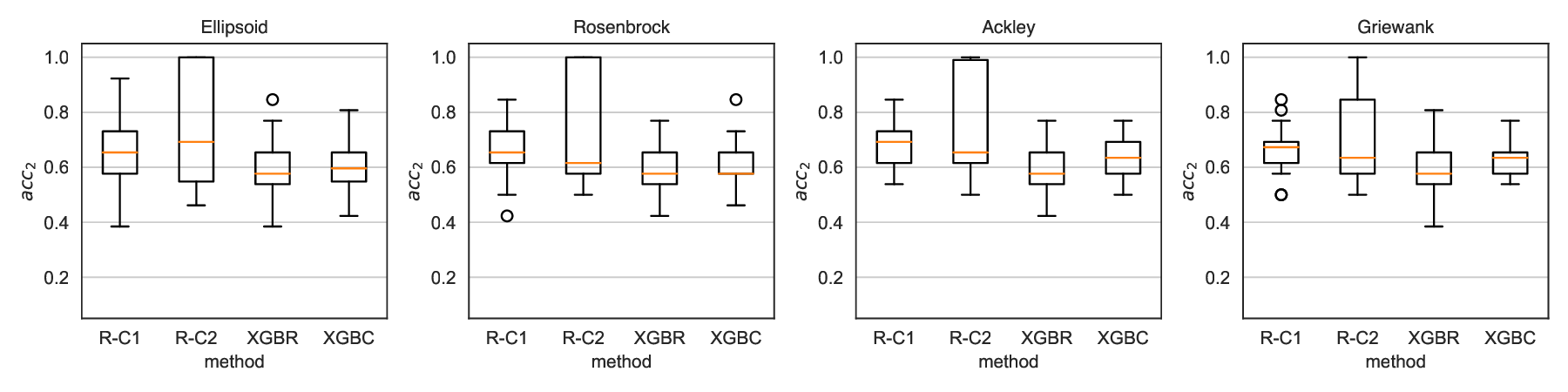} \\
\caption{Box chart of the $acc_2$ statistics of $\mathcal{P}_u$ selections for different surrogate models.}
\label{fig:select_unevaluated}
\end{figure}
Figure~\ref{fig:select_unevaluated} shows the ability of different models to select $\mathcal{P}_u$, with $t=50\%$. It can be observed that the R-C1 and R-C2 criteria exhibit better performance than XGBC and XGBR. Among them, the Interquartile range of R-C1 is more concentrated, but there are some outliers, while the maximum value range of R-C2 is more optimal.

Based on the analysis above, it can be concluded that the relation model can provide more accurate and detailed partitioning than classification models, while avoiding overfitting of data in regression models and losing the order of ranks between test points. Therefore, it is more suitable for use in model-assisted scenarios.

\subsection{Importance of unevaluated solutions}

Another key aspect to consider is whether the algorithm's efficiency is truly improved by the unevaluated solutions. In order to investigate this, we conducted an ablation experiment by designing a variant DRSO' which expunged the population $\mathcal{P}_{u}$. Moreover, we designed variants DRSO-10\%, DRSO-30\%, and DRSO-50\%, corresponding to varied values of $t$ to examine the influence of the parameter $t$ on the algorithm's efficacy.

\begin{figure}[htbp]
  \centering
  \includegraphics[width=\textwidth,trim=0 0 0 0,clip]{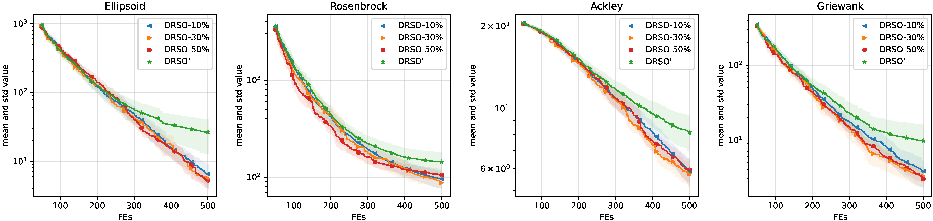} 
\caption{The runtime performance of DRSO and its variant on LZG test suite over 30 independent runs.}
\label{fig_runtime_UE}
\end{figure}

Figure~\ref{fig_runtime_UE} depicts the runtime curve of DRSO under various parameters on the LZG test suite, with a search space dimension of 20. Other parameters remain consistent with Section~\ref{sec:settings}. The figure reveals that the DRSO', lacking unevaluated solutions, converges at a slower pace than DRSO, indicating that high-quality unevaluated solutions can effectively enhance the algorithm's convergence speed. With regards to the performance of DRSO under different values of $t$, it can be observed that DRSO performs better under $t=30\%$ and $50\%$ than under $t=10\%$ on the Ellipsoid and Griewank test functions. On the Ackley function, the performance of all three values of $t$ is comparable. On the Rosenbrock function, the performance of $t=10\%$ is intermediate to that of $t=30\%$ and $t=50\%$. We surmise that when $t=10\%$, the algorithm is excessively focused on the optimal region, leading to inadequate diversity provided by unevaluated solutions in population $\mathcal{P}$ and resulting in a decrease in performance. Therefore, we recommend $t=50\%$ as the default parameter for the algorithm in this context.

\section{Conclusions}
\label{sec:conclusions}
This paper highlights an objective but often overlooked issue in SAEAs, which is the lack of variance in the adjacent population $\mathcal{P}$ due to the limited number of solutions selected for real evaluation in each iteration. To tackle this problem, this work proposes a simple method of generating new solutions based on unevaluated solutions. Employing surrogate models, the best solution in the offspring population is selected for real evaluation, and the archive and surrogate model are updated accordingly. Additionally, some potential `good' solution solutions are directly used to generate offspring without evaluation. We have designed customized relation-based surrogate models for SAEAs. Two specific relation model construction methods namely the fitness criterion (C1) and category criterion (C2), are proposed to address two selection scenarios. The C1 criterion constructs a relation pair based on relative fitness, while the C2 criterion divides the data into categories and constructs relation pairs based on category. XGBoost is utilized for data learning, and `voting-scoring' strategies are designed to enhance the model's predictive ability. Some reproduction methods in EDA/LS are employed to generate offspring solutions, and unevaluated solutions are utilized to update the VWH model. Ultimately, a dual relation models-assisted single-objective optimization algorithm (DRSO) is designed.

To verify the effectiveness of the relation model, and to demonstrate the search capability of the DRSO, This work conducted experiments on the LZG and YLL test suites in 20 and 50-dimensional search spaces. The DRSO algorithm was compared with EAs, SAEAs, and BOs, and it showed strong competitiveness. Through ablation experiments, the efficacy of each module was verified. Furthermore, the paper also scrutinized the fitting ability of the relation model to the function landscape and the predictive ability for new solution quality. The effectiveness of unevaluated solutions in the algorithm search process was affirmed through experiments and analysis of algorithm hyperparameters. Overall, the results of the experiments testify to the effectiveness of the relation model and the competitiveness of the proposed DRSO with unevaluated solutions.

In future work, it is worth exploring more detailed strategies for using unevaluated solutions to improve the quality of new solutions. The relation model can be tried on more algorithm frameworks and problem types.

\bibliographystyle{IEEEtran}
\bibliography{mybibliography}

\begin{thebibliography}{10}
\providecommand{\url}[1]{#1}
\csname url@samestyle\endcsname
\providecommand{\newblock}{\relax}
\providecommand{\bibinfo}[2]{#2}
\providecommand{\BIBentrySTDinterwordspacing}{\spaceskip=0pt\relax}
\providecommand{\BIBentryALTinterwordstretchfactor}{4}
\providecommand{\BIBentryALTinterwordspacing}{\spaceskip=\fontdimen2\font plus
\BIBentryALTinterwordstretchfactor\fontdimen3\font minus
  \fontdimen4\font\relax}
\providecommand{\BIBforeignlanguage}[2]{{%
\expandafter\ifx\csname l@#1\endcsname\relax
\typeout{** WARNING: IEEEtran.bst: No hyphenation pattern has been}%
\typeout{** loaded for the language `#1'. Using the pattern for}%
\typeout{** the default language instead.}%
\else
\language=\csname l@#1\endcsname
\fi
#2}}
\providecommand{\BIBdecl}{\relax}
\BIBdecl

\bibitem{back1997handbook}
T.~B{\"a}ck, D.~B. Fogel, and Z.~Michalewicz, ``Handbook of evolutionary
  computation,'' \emph{Release}, vol.~97, no.~1, p.~B1, 1997.

\bibitem{jinSurrogateassistedEvolutionaryComputation2011a}
Y.~Jin, ``Surrogate-assisted evolutionary computation: Recent advances and
  future challenges,'' \emph{Swarm and Evolutionary Computation}, vol.~1,
  no.~2, pp. 61--70, 2011.

\bibitem{DBLP:journals/chinaf/LiuHQQY22}
\BIBentryALTinterwordspacing
Y.~Liu, Y.~Hu, H.~Qian, C.~Qian, and Y.~Yu, ``Zoopt: a toolbox for
  derivative-free optimization,'' \emph{Sci. China Inf. Sci.}, vol.~65, no.~10,
  2022. [Online]. Available: \url{https://doi.org/10.1007/s11432-021-3416-y}
\BIBentrySTDinterwordspacing

\bibitem{LiEvolutionaryComputationExpensive2022}
J.-Y. Li, Z.-H. Zhan, and J.~Zhang, ``Evolutionary computation for expensive
  optimization: A survey,'' \emph{Machine Intelligence Research}, vol.~19,
  no.~1, pp. 3--23, 2022.

\bibitem{zhan2020expected}
D.~Zhan and H.~Xing, ``Expected improvement for expensive optimization: a
  review,'' \emph{Journal of Global Optimization}, vol.~78, no.~3, pp.
  507--544, 2020.

\bibitem{zhang2018preselection}
J.~Zhang, A.~Zhou, and G.~Zhang, ``Preselection via classification: a case
  study on global optimisation,'' \emph{International Journal of Bio-Inspired
  Computation}, vol.~11, no.~4, pp. 267--281, 2018.

\bibitem{haoBinaryRelationLearning2020}
H.~Hao, J.~Zhang, X.~Lu, and A.~Zhou, ``Binary relation learning and
  classifying for preselection in evolutionary algorithms,'' \emph{IEEE
  Transactions on Evolutionary Computation}, vol.~24, no.~6, pp. 1125--1139,
  2020-12.

\bibitem{journals/tec/LiuZG14}
B.~Liu, Q.~Zhang, and G.~G.~E. Gielen, ``A gaussian process surrogate model
  assisted evolutionary algorithm for medium scale expensive optimization
  problems,'' \emph{{IEEE} Trans. Evol. Comput.}, vol.~18, no.~2, pp. 180--192,
  2014.

\bibitem{chen2020surrogate}
G.~Chen, K.~Zhang, X.~Xue, L.~Zhang, J.~Yao, H.~Sun, L.~Fan, and Y.~Yang,
  ``Surrogate-assisted evolutionary algorithm with dimensionality reduction
  method for water flooding production optimization,'' \emph{Journal of
  Petroleum Science and Engineering}, vol. 185, p. 106633, 2020.

\bibitem{journals/tec/SunJCDZ17}
C.~Sun, Y.~Jin, R.~Cheng, J.~Ding, and J.~Zeng, ``Surrogate-assisted
  cooperative swarm optimization of high-dimensional expensive problems,''
  \emph{{IEEE} Trans. Evol. Comput.}, vol.~21, no.~4, pp. 644--660, 2017.

\bibitem{willmes2003comparing}
L.~Willmes, T.~Back, Y.~Jin, and B.~Sendhoff, ``Comparing neural networks and
  kriging for fitness approximation in evolutionary optimization,'' in
  \emph{The 2003 Congress on Evolutionary Computation, 2003. CEC'03.},
  vol.~1.\hskip 1em plus 0.5em minus 0.4em\relax IEEE, 2003, pp. 663--670.

\bibitem{li2019ibea}
H.-r. Li, F.-z. He, and X.-h. Yan, ``Ibea-svm: an indicator-based evolutionary
  algorithm based on pre-selection with classification guided by svm,''
  \emph{Applied Mathematics-A Journal of Chinese Universities}, vol.~34, no.~1,
  pp. 1--26, 2019.

\bibitem{journals/tec/PanHTWZJ19}
L.~Pan, C.~He, Y.~Tian, H.~Wang, X.~Zhang, and Y.~Jin, ``A classification-based
  surrogate-assisted evolutionary algorithm for expensive many-objective
  optimization,'' \emph{{IEEE} Trans. Evol. Comput.}, vol.~23, no.~1, pp.
  74--88, 2019.

\bibitem{loshchilov2010comparison}
I.~Loshchilov, M.~Schoenauer, and M.~Sebag, ``Comparison-based optimizers need
  comparison-based surrogates,'' in \emph{International Conference on Parallel
  Problem Solving from Nature}.\hskip 1em plus 0.5em minus 0.4em\relax
  Springer, 2010, pp. 364--373.

\bibitem{hao2022expensive}
H.~Hao, A.~Zhou, H.~Qian, and H.~Zhang, ``Expensive multiobjective optimization
  by relation learning and prediction,'' \emph{IEEE Transactions on
  Evolutionary Computation}, 2022.

\bibitem{hao2021approximated}
H.~Hao, A.~Zhou, and H.~Zhang, ``An approximated domination relationship based
  on binary classifiers for evolutionary multiobjective optimization,'' in
  \emph{2021 IEEE Congress on Evolutionary Computation (CEC)}.\hskip 1em plus
  0.5em minus 0.4em\relax IEEE, 2021, pp. 2427--2434.

\bibitem{hao2023relation}
H.~Hao and A.~Zhou, ``A relation surrogate model for expensive multiobjective
  continuous and combinatorial optimization,'' in \emph{Evolutionary
  Multi-Criterion Optimization: 12th International Conference, EMO 2023,
  Leiden, The Netherlands, March 20--24, 2023, Proceedings}.\hskip 1em plus
  0.5em minus 0.4em\relax Springer, 2023, pp. 205--217.

\bibitem{yuan2021expensive}
Y.~Yuan and W.~Banzhaf, ``Expensive multiobjective evolutionary optimization
  assisted by dominance prediction,'' \emph{IEEE Transactions on Evolutionary
  Computation}, vol.~26, no.~1, pp. 159--173, 2021.

\bibitem{tian2023pairwise}
Y.~Tian, J.~Hu, C.~He, H.~Ma, L.~Zhang, and X.~Zhang, ``A pairwise comparison
  based surrogate-assisted evolutionary algorithm for expensive multi-objective
  optimization,'' \emph{Swarm and Evolutionary Computation}, vol.~80, p.
  101323, 2023.

\bibitem{zhang2022dual}
J.~Zhang, L.~He, and H.~Ishibuchi, ``Dual fuzzy classifier-based evolutionary
  algorithm for expensive multiobjective optimization,'' \emph{IEEE
  Transactions on Evolutionary Computation}, 2022.

\bibitem{zhou2019fuzzy}
A.~Zhou, J.~Zhang, J.~Sun, and G.~Zhang, ``Fuzzy-classification assisted
  solution preselection in evolutionary optimization,'' in \emph{Proceedings of
  the AAAI Conference on Artificial Intelligence}, vol.~33, no.~01, 2019, pp.
  2403--2410.

\bibitem{LiExpensiveOptimizationSurrogateAssisted2022a}
G.~Li, Z.~Wang, and M.~Gong, ``Expensive optimization via surrogate-assisted
  and model-free evolutionary optimization,'' \emph{IEEE Transactions on
  Systems, Man, and Cybernetics: Systems}, pp. 1--12, 2022.

\bibitem{srinivas2009gaussian}
N.~Srinivas, A.~Krause, S.~M. Kakade, and M.~Seeger, ``Gaussian process
  optimization in the bandit setting: No regret and experimental design,''
  \emph{arXiv preprint arXiv:0912.3995}, 2009.

\bibitem{jones1998efficient}
D.~R. Jones, M.~Schonlau, and W.~J. Welch, ``Efficient global optimization of
  expensive black-box functions,'' \emph{Journal of Global optimization},
  vol.~13, no.~4, p. 455, 1998.

\bibitem{mockus1998application}
J.~Mockus, ``The application of bayesian methods for seeking the extremum,''
  \emph{Towards global optimization}, vol.~2, p. 117, 1998.

\bibitem{hao2022surrogate}
H.~Hao, S.~Wang, B.~Li, and A.~Zhou, ``A surrogate model assisted estimation of
  distribution algorithm with mutil-acquisition functions for expensive
  optimization,'' in \emph{2022 IEEE Congress on Evolutionary Computation
  (CEC)}.\hskip 1em plus 0.5em minus 0.4em\relax IEEE, 2022, pp. 1--8.

\bibitem{DBLP:conf/uai/HoffmanBF11}
M.~Hoffman, E.~Brochu, and N.~de~Freitas, ``Portfolio allocation for bayesian
  optimization,'' in \emph{{UAI} 2011, Proceedings of the Twenty-Seventh
  Conference on Uncertainty in Artificial Intelligence, Barcelona, Spain, July
  14-17, 2011}, F.~G. Cozman and A.~Pfeffer, Eds.\hskip 1em plus 0.5em minus
  0.4em\relax {AUAI} Press, 2011, pp. 327--336.

\bibitem{journals/tcyb/LiCGS21}
F.~Li, X.~Cai, L.~Gao, and W.~Shen, ``A surrogate-assisted multiswarm
  optimization algorithm for high-dimensional computationally expensive
  problems,'' \emph{{IEEE} Trans. Cybern.}, vol.~51, no.~3, pp. 1390--1402,
  2021.

\bibitem{sun2017surrogate}
C.~Sun, Y.~Jin, R.~Cheng, J.~Ding, and J.~Zeng, ``Surrogate-assisted
  cooperative swarm optimization of high-dimensional expensive problems,''
  \emph{IEEE Transactions on Evolutionary Computation}, vol.~21, no.~4, pp.
  644--660, 2017.

\bibitem{wei2020classifier}
F.-F. Wei, W.-N. Chen, Q.~Yang, J.~Deng, X.-N. Luo, H.~Jin, and J.~Zhang, ``A
  classifier-assisted level-based learning swarm optimizer for expensive
  optimization,'' \emph{IEEE Transactions on Evolutionary Computation},
  vol.~25, no.~2, pp. 219--233, 2020.

\bibitem{lian2005multiobjective}
Y.~Lian and M.-S. Liou, ``Multiobjective optimization using coupled response
  surface model and evolutionary algorithm.'' \emph{AIAA journal}, vol.~43,
  no.~6, pp. 1316--1325, 2005.

\bibitem{conf/aaai/ZhouZSZ19}
A.~Zhou, J.~Zhang, J.~Sun, and G.~Zhang, ``Fuzzy-classification assisted
  solution preselection in evolutionary optimization,'' in \emph{The
  Thirty-Third {AAAI} Conference on Artificial Intelligence}, 2019, pp.
  2403--2410.

\bibitem{zhang2007moea}
Q.~Zhang and H.~Li, ``Moea/d: A multiobjective evolutionary algorithm based on
  decomposition,'' \emph{IEEE Transactions on evolutionary computation},
  vol.~11, no.~6, pp. 712--731, 2007.

\bibitem{holland1992adaptation}
J.~H. Holland, \emph{Adaptation in natural and artificial systems: an
  introductory analysis with applications to biology, control, and artificial
  intelligence}.\hskip 1em plus 0.5em minus 0.4em\relax MIT press, 1992.

\bibitem{storn1997differential}
R.~Storn and K.~Price, ``Differential evolution-a simple and efficient
  heuristic for global optimization over continuous spaces,'' \emph{Journal of
  global optimization}, vol.~11, no.~4, p. 341, 1997.

\bibitem{journals/tec/ZhouSZ15}
A.~Zhou, J.~Sun, and Q.~Zhang, ``An estimation of distribution algorithm with
  cheap and expensive local search methods,'' \emph{{IEEE} Trans. Evol.
  Comput.}, vol.~19, no.~6, pp. 807--822, 2015.

\bibitem{hollander2013nonparametric}
M.~Hollander, D.~A. Wolfe, and E.~Chicken, \emph{Nonparametric Statistical
  Methods}.\hskip 1em plus 0.5em minus 0.4em\relax John Wiley \& Sons, 2013,
  vol. 751.

\bibitem{mckay2000comparisona}
M.~D. Mckay, R.~J. Beckman, and W.~J. Conover, ``A {{Comparison}} of {{Three
  Methods}} for {{Selecting Values}} of {{Input Variables}} in the {{Analysis}}
  of {{Output From}} a {{Computer Code}},'' \emph{Technometrics}, vol.~42,
  no.~1, pp. 55--61, 2000-02.

\bibitem{chen2016xgboost}
T.~Chen and C.~Guestrin, ``Xgboost: A scalable tree boosting system,'' in
  \emph{Proceedings of the 22nd acm sigkdd international conference on
  knowledge discovery and data mining}, 2016, pp. 785--794.

\bibitem{yao1999evolutionary}
X.~Yao, Y.~Liu, and G.~Lin, ``Evolutionary programming made faster,''
  \emph{IEEE Transactions on Evolutionary Computation}, vol.~3, no.~2, pp.
  82--102, 1999.

\bibitem{hansen2001completely}
N.~Hansen and A.~Ostermeier, ``Completely derandomized self-adaptation in
  evolution strategies,'' \emph{Evolutionary computation}, vol.~9, no.~2, pp.
  159--195, 2001.

\bibitem{tian2017platemo}
Y.~Tian, R.~Cheng, X.~Zhang, and Y.~Jin, ``{{PlatEMO}}: {{A MATLAB Platform}}
  for {{Evolutionary Multi}}-{{Objective Optimization}} [{{Educational
  Forum}}],'' \emph{IEEE Computational Intelligence Magazine}, vol.~12, no.~4,
  pp. 73--87, 2017-11.

\bibitem{friedman1937use}
M.~Friedman, ``The use of ranks to avoid the assumption of normality implicit
  in the analysis of variance,'' \emph{Journal of the american statistical
  association}, vol.~32, no. 200, pp. 675--701, 1937.

\end{thebibliography}



\end{document}